\PassOptionsToPackage{numbers}{natbib}

\documentclass{article}

\usepackage[preprint]{neurips}

\usepackage[utf8]{inputenc} 
\usepackage[T1]{fontenc}    
\usepackage{hyperref}       
\usepackage{xurl}            
\usepackage{booktabs}       
\usepackage{amsfonts}       
\usepackage{amssymb}        
\usepackage{amsmath}        
\usepackage{nicefrac}       
\usepackage{microtype}      
\usepackage{xcolor}         
\usepackage{graphicx}       
\hypersetup{hidelinks}
\usepackage{multirow}
\usepackage{subcaption}
\usepackage{float}          

\usepackage{tikz}
\usetikzlibrary{
    arrows.meta,
    positioning,
    calc,
    backgrounds,
    fit
}

\usepackage{pgfplots}
\pgfplotsset{compat=1.18}
\usepgfplotslibrary{fillbetween}

\usepackage{siunitx}

\definecolor{rgbblue}{HTML}{000000}
\definecolor{yamber}{HTML}{C47C00}
\definecolor{solvergreen}{HTML}{177E70}
\definecolor{darktext}{HTML}{263442}
\definecolor{softgray}{HTML}{E8EDF1}
\definecolor{warningred}{HTML}{C4473A}

\usepackage{cleveref}

\title{What Builds the Scene? Luminance Dominates Geometry Formation in 3D Gaussian Splatting}

\author{
    Rezvan Joshaghani \qquad Steven Cutchin \\
    Boise State University \\
    \texttt{RezvanJoshaghani@u.boisestate.edu} \qquad
    \texttt{StevenCutchin@boisestate.edu}
}

\begin{document}

\maketitle

\pgfplotstableread[col sep=tab]{figures/data/delayed_chroma_garden.tsv}\chromadata

\begin{abstract}
Standard 3D Gaussian Splatting (3DGS) learns geometry and appearance jointly from RGB supervision, making it difficult to isolate how luminance and chroma contribute to the learned representation. We study this by training models under different channel supervision, freezing their non-appearance parameters (position, scale, rotation, and opacity), and re-estimating appearance with the same solver before comparing held-out reconstruction. Across eleven benchmark scenes with four independent runs each, geometry learned from luminance alone supports held-out reconstruction $0.085$ dB below RGB-trained geometry on average. If chroma is deleted from a trained model, a sufficiently expressive solver can re-fit it on the frozen geometry to the original quality or slightly better. Higher-order spherical harmonics contribute much more reconstruction quality to luminance than to chroma, improving PSNR by 1.44 dB versus 0.19 dB on average, although on mirror-like surfaces hue does still change with viewpoint. The luminance advantage is even larger when geometry is being formed. Chroma-only supervision produces geometry 3.9--5.5 dB worse than luminance-only supervision after the same appearance solve; densification explains part of this gap. Overall, geometry formation in standard 3DGS is strongly luminance-dominated but not luminance-exclusive, and much of the chromatic appearance can be recovered after spatial support has formed.
\end{abstract}

\section{Introduction}
\label{sec:intro}

Standard RGB optimization in 3D Gaussian Splatting (3DGS)~\cite{kerbl20233d}
combines two roles of the image signal: \emph{building the scene} and
\emph{coloring the scene}. A standard degree-3 model represents appearance
with 16 spherical-harmonic (SH) coefficients per color channel, or 48
appearance coefficients per Gaussian. Because SH evaluation is linear in its
coefficients, these appearance parameters can be re-expressed under a linear
change of color basis. For example, YCbCr separates the image signal into one
luminance-like channel, \(Y\), and two chromatic channels, \(Cb\) and \(Cr\).
Yet these appearance parameters are not learned independently of scene
structure. Standard 3DGS jointly optimizes Gaussian positions, scales,
rotations, opacities, and SH coefficients against the same image-space
reconstruction loss, while densification and pruning are driven by statistics
of that optimization. An RGB residual therefore does not reveal which part of
the color signal caused the representation to form.

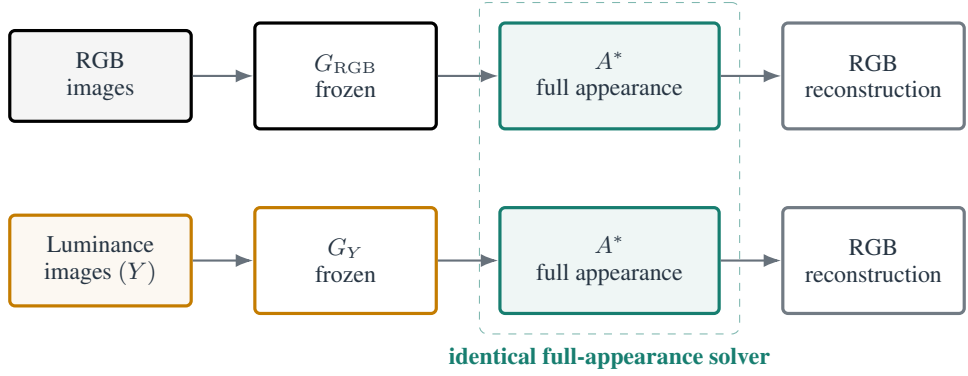
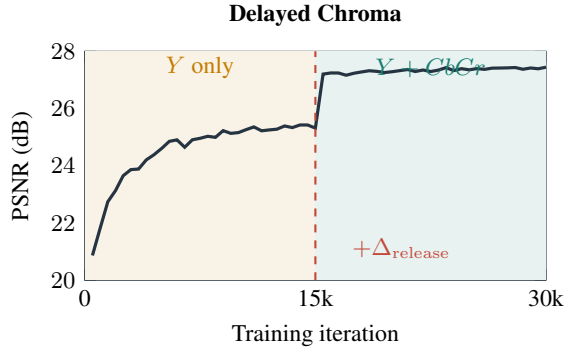
\begin{figure}[h!]
    \centering

    \begin{subfigure}[t]{\textwidth}
    \centering

    \begin{tikzpicture}[
        >=Latex,
        node distance=7mm and 8mm,
        every node/.style={
            font=\small,
            text=darktext,
            align=center
        },
        rgbinput/.style={
            draw=rgbblue,
            very thick,
            rounded corners=2pt,
            minimum width=24mm,
            minimum height=12mm,
            fill=rgbblue!4
        },
        yinput/.style={
            draw=yamber,
            very thick,
            rounded corners=2pt,
            minimum width=24mm,
            minimum height=12mm,
            fill=yamber!5
        },
        rgbgeom/.style={
            draw=rgbblue,
            very thick,
            rounded corners=2pt,
            minimum width=24mm,
            minimum height=14mm
        },
        ygeom/.style={
            draw=yamber,
            very thick,
            rounded corners=2pt,
            minimum width=24mm,
            minimum height=14mm
        },
        solver/.style={
            draw=solvergreen,
            very thick,
            rounded corners=2pt,
            minimum width=29mm,
            minimum height=14mm,
            fill=solvergreen!7
        },
        output/.style={
            draw=darktext!65,
            very thick,
            rounded corners=2pt,
            minimum width=24mm,
            minimum height=14mm
        },
        flow/.style={
            ->,
            thick,
            draw=darktext!70
        }
    ]

    \node[rgbinput] (rgbin) {RGB\\images};
    \node[rgbgeom, right=of rgbin] (grgb) {$G_{\mathrm{RGB}}$\\frozen};
    \node[solver, right=of grgb] (argb) {$A^*$\\full appearance};
    \node[output, right=of argb] (outrgb) {RGB\\reconstruction};

    \draw[flow] (rgbin) -- (grgb);
    \draw[flow] (grgb) -- (argb);
    \draw[flow] (argb) -- (outrgb);

    \node[yinput, below=12mm of rgbin] (yin) {Luminance\\images $(Y)$};
    \node[ygeom, right=of yin] (gy) {$G_Y$\\frozen};
    \node[solver, right=of gy] (ay) {$A^*$\\full appearance};
    \node[output, right=of ay] (outy) {RGB\\reconstruction};

    \draw[flow] (yin) -- (gy);
    \draw[flow] (gy) -- (ay);
    \draw[flow] (ay) -- (outy);

    \begin{scope}[on background layer]
        \node[
            draw=solvergreen!55,
            dashed,
            rounded corners=3pt,
            fit=(argb)(ay),
            inner sep=2.5mm
        ] {};
    \end{scope}

    \node[
        font=\footnotesize\bfseries,
        text=solvergreen,
        below=3mm of ay
    ]
    {identical full-appearance solver};

    \end{tikzpicture}

    \caption{Different geometry supervision, identical appearance fitting.}
    \label{fig:two_arm_setup}
    \end{subfigure}

    \vspace{1.5mm}

    \begin{subfigure}[t]{0.55\textwidth}
    \centering

    \begin{tikzpicture}
    \begin{axis}[
        width=\linewidth,
        height=0.6\linewidth,
        xmin=0,
        xmax=30000,
        ymin=20,
        ymax=28,
        xlabel={Training iteration},
        ylabel={PSNR (dB)},
        xtick={0,15000,30000},
        xticklabels={0,15k,30k},
        scaled x ticks=false, 
        tick label style={font=\small},
        label style={font=\small},
        title={\textbf{Delayed Chroma}},
        title style={font=\small},
        axis line style={draw=darktext},
        clip=true
    ]

    \fill[yamber!10]
        (rel axis cs:0,0)
        rectangle
        (rel axis cs:0.5,1);

    \fill[solvergreen!10]
        (rel axis cs:0.5,0)
        rectangle
        (rel axis cs:1,1);

    \addplot[
        very thick,
        darktext
    ]
    table[
        x=iteration,
        y=psnr,
        col sep=tab
    ] {figures/data/delayed_chroma_garden.tsv};

    \draw[
        warningred,
        dashed,
        thick
    ]
    (rel axis cs:0.5,0)
    --
    (rel axis cs:0.5,1);

    \node[
        font=\small,
        text=yamber
    ]
    at (rel axis cs:0.25,0.93)
    {$Y$ only};

    \node[
        font=\small,
        text=solvergreen
    ]
    at (rel axis cs:0.75,0.93)
    {$Y+CbCr$};

    \node[
        font=\small,
        text=warningred,
        anchor=west
    ]
    at (rel axis cs:0.56,0.14)
    {$+\Delta_{\mathrm{release}}$};

    \end{axis}
    \end{tikzpicture}

    \caption{Motivating Delayed-Chroma training observation.}
    \label{fig:delayed_chroma_hook}
    \end{subfigure}

    \caption{
    \textbf{Separating geometry supervision from appearance fitting.}
    (\textbf{a}) One geometry is trained with full RGB supervision and the
    other with luminance-only supervision; after freezing geometry, the same
    full-appearance solver $A^*$ is applied to both.
    (\textbf{b}) The Delayed-Chroma observation that motivated the study:
    chromatic supervision is withheld during an initial luminance phase and
    introduced later. This training behavior motivates the geometry test but
    is not itself evidence of geometry equivalence.
    }
    \label{fig:experimental_overview}

\end{figure}

A simple training intervention exposes this ambiguity. We withhold chromatic
supervision during the first half of training and optimize using only \(Y\);
\(Cb\) and \(Cr\) supervision are introduced later
(Fig.~\ref{fig:experimental_overview}b). When chroma is released, held-out
reconstruction quality rises sharply, yet the delayed schedules often converge
near ordinary RGB training. This does not tell us why. Luminance may already
have built the geometry, with chroma filling in appearance; alternatively,
the later full-color phase may repair geometry left incomplete by luminance. A successful final reconstruction after delayed chroma
therefore does not tell us which signal actually built the representation.

Several results already point to a separation of these roles along the
luminance--chroma axis. Monochromatic reconstruction followed by colorization
shows that a useful spatial representation can be learned without observed
chroma and colored afterward
\citep{cheng2024colorizing,dhiman2025chromadistill,color3d2026}. Instant
Colorization shows that appearance can be re-estimated on a fixed Gaussian
representation without further geometric optimization
\citep{lieber2026instant}. LCD-GS and palette-based editing separate
luminance-like and chromatic components explicitly, for HDR reconstruction
and for appearance manipulation \citep{lcdgs2026,chao2026palette}. GSICO,
HiGS, and YUV-domain channel decorrelation show that higher-order chromatic
SH can be pruned or coarsely quantized at little cost
\citep{gsico2026,higs2026,sony2026decorrelation}. Outside neural rendering,
image coding has long subsampled chroma and geometric vision runs on
intensity alone \citep{jpeg_chroma_subsampling,sift,colmap}. Together these
results establish feasibility: structure can exist without observed chroma,
color can be fitted after geometry exists, and chromatic appearance can be
stored compactly.

Natural-image statistics supply a prior for what to expect. In natural
images, luminance carries most of the spatial detail and contrast, while the
chromatic channels are smoother and lower in amplitude
\citep{ruderman1998statistics,mullen1985contrast}. That prior predicts that
luminance should matter more for geometry. It does not tell us how much worse
chroma-only training will perform, or how much of that gap comes from
densification, the process that adds Gaussians during training. We test whether
the gap persists when both conditions keep the same set of Gaussians. We also
measure how much chroma can be fitted on fixed geometry and how much each
channel group benefits from view-dependent SH terms.

These prior methods do not establish what chromatic supervision contributes to geometry learned during RGB training. Nor do they measure how changing when observed chroma enters the loss affects the learned geometry. The Delayed-Chroma result shows that final reconstruction quality alone cannot distinguish the possible explanations. To separate them, we freeze Gaussian position, scale, rotation, and opacity after training and apply the same full degree-3 appearance solve \(A^*\) to each geometry (Fig.~\ref{fig:two_arm_setup}; Sec.~\ref{sec:luminance-geometry}). This measures the reconstruction each geometry can support after appearance is re-estimated under the same conditions. Throughout, geometry means the non-appearance parameters of the model, including opacity, and every model starts from the SfM initialization provided with its dataset. Our claims therefore concern reconstruction of the captured benchmark images after standard 3DGS optimization and densification from that initialization; they do not concern the physical accuracy of its surfaces, and they do not concern geometry formed without a spatial prior. We combine this comparison with fixed-geometry chroma recovery and the SH-capacity sweep (Sec.~\ref{sec:appearance-capacity}), and with direct changes to channel supervision (Sec.~\ref{sec:channel-geometry}). All experiments use BT.601 YCbCr and the eleven-scene 3DGS benchmark unless stated otherwise.

\paragraph{1.~Geometry trained on luminance alone supports nearly the same reconstruction as RGB-trained geometry.}
After the same appearance solve, validated in contribution 4, with four independent runs per condition on every scene, the average difference in held-out PSNR is $-0.085\pm0.037$ dB, a small advantage for RGB (95\% CI $[-0.166,-0.003]$). Eight of eleven scenes fall inside the $\pm0.15$ dB band we set before the comparison. The three outliers are \textsc{kitchen}, \textsc{dr.~johnson}, and \textsc{playroom}, where RGB-trained geometry keeps an average advantage of $0.2$--$0.3$ dB over four independent runs. At seed 0, SSIM moves in the same direction by $0.001$, and LPIPS slightly favors luminance-trained geometry on average while favoring RGB-trained geometry on two of the three departure scenes. Controls for Gaussian count, appearance basis, and solver iteration budget support this interpretation in the tested settings.

\paragraph{2.~Luminance provides most of the signal for learning geometry.}
On three scenes, geometry trained on chroma alone supports $3.9$--$5.5$ dB less reconstruction than geometry trained on luminance alone. Part of this gap comes from densification, because chroma produces smaller positional gradients and therefore fewer new Gaussians. However, when both conditions start from the same Gaussians with densification turned off, chroma-only geometry is still $1.4$--$2.5$ dB worse. Chroma does still affect geometry. Giving each training view the chroma of another view, while leaving luminance unchanged, lowers reconstruction by $0.15$--$0.67$ dB, and the drop appears in all four adversarial seeds.
This effect is consistent but much smaller than the effect of removing luminance.

\paragraph{3.~Higher-order SH matters far more for luminance than for chroma.}
Removing SH degrees 1--3 lowers held-out PSNR by $1.441$ dB for luminance but only $0.187$ dB for chroma on average, and luminance loses more on every benchmark scene. In linear terms, held-out MSE rises by $35\%$ for luminance and $3.6\%$ for chroma. The same ordering holds on specular scenes. This does not mean that chroma is view independent. On the chrome ball, measured directly from the ground-truth views, chromaticity changes substantially with viewpoint.

\paragraph{4.~Chroma can be fully re-estimated on fixed geometry.}
We remove the chroma from trained RGB models and re-estimate it from the training images with the geometry frozen. Recovery rises from $0.300$ with a single scene-wide tint to $1.017$ with coupled degree-3 fitting, where $1$ matches the trained model. Most of the chroma loss seen with simpler solves therefore comes from the solver, not from the geometry. Because this solve recovers appearance fully, we use it to compare geometries trained under different supervision without carrying over the appearance each model learned.

\paragraph{5.~Delayed chroma follows the same asymmetry.}
Withholding chroma for the first half of training preserves final quality on two of three replicated scenes and costs $0.27$ dB on \textsc{bonsai}. When chroma is introduced, quality rises sharply, and the rise moves with the release time rather than with other training events such as the end of densification. When training continues from the same luminance-only checkpoint, adding chroma improves the reconstruction the final geometry supports by about $0.1$ dB, which matches the RGB advantage measured in the main comparison. The pattern largely holds in an independent 3DGS implementation and in BT.709, YCoCg, and YIQ.

Together, these results show that luminance provides most of the signal for learning geometry that supports reconstruction, and most of the benefit of view-dependent appearance. Chroma plays a smaller role in forming geometry, and most of its appearance can be fitted afterward with little higher-order SH. However, the asymmetry is not absolute. Chroma can still change the learned geometry, RGB-trained geometry keeps an advantage on three scenes, and view-dependent chromatic reflections can require higher-order chromatic SH.

\section{Related Work}
\label{sec:related}

The distinction between luminance and chroma has appeared in several areas of 3D reconstruction and Gaussian Splatting. It has been used for monochromatic reconstruction and colorization, appearance editing, progressive training, and compression. These works provide useful evidence about what can be learned from luminance alone and how chromatic appearance can be represented after training. They do not directly measure what chroma contributes while standard RGB-supervised 3DGS learns its geometry.

\subsection{Monochromatic Reconstruction and 3D Colorization}

Several methods reconstruct scenes from monochromatic views and add color afterward. \citet{cheng2024colorizing} learn luminance and density from monochromatic multiview images, then predict color for the reconstructed radiance field. ChromaDistill \citep{dhiman2025chromadistill} studies a similar problem primarily with Plenoxels and also reports a 3DGS variant in its supplementary material. In that variant, the model is trained with luminance alone, and chroma is introduced through knowledge distillation from an image colorization model. Color3D \citep{color3d2026} extends this to both static and dynamic scenes. It combines an image colorizer with a Lab-space Gaussian representation in which luminance and chroma are optimized separately. In all three cases, the input images contain no chromatic information. Color is estimated afterward using an image colorization model.

Our setting removes chroma at a different point. Monochromatic reconstruction starts from input images with no chromatic observations. We start from the same RGB images used in standard 3DGS training. One model is trained on luminance alone and the other receives full RGB supervision. We then compare the learned geometries after the same full-appearance solve. This measures what chromatic supervision contributes while geometry is being learned.

\subsection{Luminance and Color-Space Treatments in Gaussian Splatting}

LCD-GS \citep{lcdgs2026} uses a luminance--chromaticity decomposition for HDR reconstruction. Luminance-GS \citep{luminancegs2025} addresses difficult illumination through per-view color mapping and view-adaptive curve correction during 3DGS training. Thinking Like Van Gogh \citep{vangoghgs2026} is a 3D stylizer that separates luminance-related structure from color to control the geometric deformation separately from appearance changes.

In these methods, the separation between luminance and color is built into the representation or optimization. They do not test whether standard 3DGS training already relies more on luminance than on chroma when both are available through RGB supervision. We keep the representation, rasterizer, and geometric parameters unchanged and modify only the supervision. The learned geometries are then compared after the same full-appearance solve.

\subsection{Appearance Recovery on Fixed Geometry}

Other work solves for appearance after the Gaussian geometry has already been learned. Instant Colorization \citep{lieber2026instant} is the closest algorithmic precedent for our post-training appearance recovery experiments. \citet{lieber2026instant} freeze a pretrained Gaussian representation and solve for new appearance coefficients using visibility-weighted least squares. Their formulation supports view-dependent spherical-harmonic appearance and applications such as colorization, relighting, and attribute transfer. \citet{zhou2026structured} uses a similar setup to determine which spherical-harmonic coefficients can be recovered from the training views and remove poorly constrained directions from the solve. \citet{chao2026palette} keep geometry fixed while reparameterizing appearance for palette-based color and luminance editing.

A good reconstruction after an appearance solve does not guarantee the physical correctness of the underlying geometry. \citet{wang2026surfaces} show that view-dependent appearance can compensate for geometric errors while the reconstructed images still look correct. We therefore use reconstruction after appearance recovery to measure what a learned geometry can support, not whether it matches the true surface.

We build on Instant Colorization and extend its formulation to arbitrary subsets of color channels in an affine color space. Solver quality matters because incomplete appearance recovery can reduce reconstruction quality even when the geometry supports a better solution. We therefore test several solvers and use the one that best reconstructs the training images from fixed geometry. The formulation and solver evaluation are described in Sec.~\ref{sec:experimental-framework}.

\subsection{Progressive Supervision in 3DGS Training}

Several methods control when different parts of the training signal become available. FreGS \citep{zhang2024fregs} applies coarse-to-fine frequency regularization to encourage stable geometry before finer detail is introduced. Opti3DGS \citep{opti3dgs2025} progressively reduces blur in the training images so that high-frequency spatial detail enters later in training. AutoOpti3DGS \citep{autoopti3dgs2025} learns this schedule with a wavelet transform rather than setting it in advance. PDF-GS \citep{pdfgs2026} filters distracting content across training phases before the final reconstruction. These methods show that the timing of supervision can affect the geometry learned during training. Opti3DGS highlights this connection by linking delayed fine-scale supervision to reduced premature densification.

Our Delayed-Chroma experiment applies this idea to color. We withhold chroma early in training and introduce it later, while luminance remains available from the start. The 3DGS variant of ChromaDistill \citep{dhiman2025chromadistill} uses the same luminance-first schedule, training on luminance and adding chroma afterward. The difference here is that our chroma is observed in the training images rather than distilled from a colorization model, so the schedule tests what observed chroma adds. The experiment is not intended as a new training strategy. We use it to test whether the learned geometry changes when chromatic supervision becomes available.

\subsection{Appearance Compression in Trained 3DGS Models}

Prior work also examines how much appearance capacity a trained 3DGS model needs. Methods reduce storage through compact or quantized Gaussian representations \citep{niedermayr2024compressed,compact3dgs2024}, adaptive SH allocation and memory-aware parameter reduction \citep{papantonakis2024memory}, and entropy coding of Gaussian attributes \citep{entropygs2025}. Much of the stored appearance capacity can be removed or represented more compactly without substantially reducing reconstruction quality. These methods focus on compression rather than why the redundancy differs between luminance and chroma or how that difference arises during training.

GSICO \citep{gsico2026}, HiGS \citep{higs2026}, and Sony's YUV-domain channel-decorrelation method \citep{sony2026decorrelation} provide more direct evidence of this difference. GSICO transforms SH coefficients from RGB to BT.601 YUV and removes chrominance AC components before image-based coding. Sony's method transforms the SH coefficients into YUV and retains the base color and luminance terms while removing higher-order chromatic coefficients. HiGS uses quantization rather than pruning. Its optional packed SH representation converts the coefficients to YCoCg-R and assigns 6 bits to luminance and 4 bits to each chromatic component, reducing per-Gaussian SH storage from 192 to 32 bytes while maintaining nearly the same rendering quality. These results show that chromatic SH can tolerate more aggressive compression than luminance. We investigate whether this post-training difference is related to the different roles played by luminance and chroma during geometry learning.

\section{Experimental Framework}
\label{sec:experimental-framework}

\subsection{Datasets and evaluation protocol}

Our primary benchmark contains eleven standard 3DGS scenes from Mip-NeRF 360~\citep{barron2022mipnerf360}, Tanks and Temples~\citep{knapitsch2017tanks}, and Deep Blending~\citep{hedman2018deep}.\footnote{The eleven scenes are \textsc{bicycle}, \textsc{bonsai}, \textsc{counter}, \textsc{garden}, \textsc{kitchen}, \textsc{room}, and \textsc{stump} from Mip-NeRF 360; \textsc{train} and \textsc{truck} from Tanks and Temples; and \textsc{dr.\ johnson} and \textsc{playroom} from Deep Blending.} We additionally use Shiny Blender~\citep{verbin2022refnerf} for the view-dependence analysis described later. Unless stated otherwise, all models are trained for 30k iterations using the camera calibration and SfM initialization provided with each dataset. We use ``stock'' to mean the reference Inria 3DGS implementation and its defaults. Images are loaded at the stock resolution setting, which caps image width at 1600 pixels: the Mip-NeRF 360 scenes are downscaled by $1.95$--$3.24\times$ from their full-resolution images, and the Tanks and Temples and Deep Blending scenes train at native resolution. These initializations come from COLMAP~\cite{colmap}, which extracts SIFT features on grayscale images, so the initial point positions are recovered from grayscale features in every condition. Resetting appearance before evaluation removes the stored initial colors, but does not rule out effects of those colors on the training trajectory. The main geometry comparison uses four independent training runs per condition on every scene. The single-seed realization that preceded the replication is the one shown in Table~\ref{tab:solver-geometry} and is summarized in the supplement. Runs with the same seed label share the camera order, but GPU rasterization is nondeterministic, so a seed label does not fully determine a run. We analyze the runs using condition means, and report a paired analysis of same-seed-label differences in the supplement as a sensitivity check. Throughout, $\pm$ denotes a standard error unless a standard deviation is stated. Image quality is measured on the benchmark test views in RGB as PSNR, SSIM~\citep{wang2004ssim}, and LPIPS~\citep{zhang2018lpips}. PSNR is the primary metric; SSIM and LPIPS are reported for the main comparisons. 

Training in the stock 3DGS implementation also withholds eight training views, evenly spaced by image name, as a validation set. The stock delayed-chroma measurements in Sec.~\ref{sec:training-dynamics} and the supplement (the release curves of Fig.~\ref{fig:training-dynamics}, release steps, early-freeze costs, and final delayed-versus-RGB differences) are scored on these eight views. The training curve in Fig.~\ref{fig:experimental_overview}b, the FastGS~\citep{ren2026fastgs} runs, and every $A^*$ comparison, including the shared-checkpoint fork of Sec.~\ref{sec:training-dynamics}, are scored on the benchmark test views.

The appearance solve fits all non-test views, including the eight validation views. On \textsc{garden}, for example, it fits 161 views and is evaluated on 24 benchmark test views.

\subsection{Color space}

To separate luminance from chroma, we transform the training images, background color, and SH appearance coefficients from RGB to full-range BT.601 YCbCr:
\begin{align}
Y &= 0.299R+0.587G+0.114B,\\
Cb &= (B-Y)/1.772+0.5,\\
Cr &= (R-Y)/1.402+0.5.
\end{align}
Here $Y$ is luma computed from the encoded RGB images, not physical luminance; we retain the term ``luminance'' for consistency. The affine offset is applied only to the SH DC term. Higher-order coefficients receive the linear part of the transform because they represent directional changes around the base color. The implementation removes the rasterizer's $+0.5$ color bias from the DC term, applies the affine transform, and restores the bias. For full-range BT.601 the offsets cancel exactly, because the $Y$ row of the matrix sums to one and the $Cb$ and $Cr$ rows sum to zero, so the net DC transform is the linear matrix alone; the two paths agree to $10^{-15}$ numerically.
The rasterizer applies a color clamp, so rendering in RGB and then converting the image to YCbCr is not exactly identical to rendering directly in YCbCr. In our validation, the difference is below $0.11$\,dB. All fixed-geometry comparisons use the same working color space, rendering path, appearance solve, and inverse transform. This standardizes evaluation, but does not rule out geometry-dependent effects of clamping.

\subsection{Training conditions}

Standard RGB training uses the usual $\mathcal L_1+$D-SSIM loss function over all three RGB channels. The $Y$ condition uses the same loss function on the luminance channel only, while the $Cb$ and $Cr$ coefficients receive zero gradient and remain unsupervised. Because the $\mathcal L_1$ term is averaged over the supervised channels, supervising one channel does not change the loss scale by a factor of three. Unsupervised coefficients stay at their SfM initialization and are copied when Gaussians are cloned or split. The common solve replaces these coefficients before evaluation, but does not undo any effects they may have had during training. Delayed Chroma also begins with luminance-only supervision, but $Cb$ and $Cr$ are introduced at a chosen release iteration. The complementary $CbCr$ condition supervises only the two chromatic channels. We call each training condition an arm. In this paper, geometry is denoted by $G$ and includes Gaussian position, scale, rotation, and opacity. Opacity is included because 3DGS treats it as a spatial-support parameter: it drives pruning and determines which primitives contribute to a pixel. Opacity also affects rendered color, so it can retain appearance information that the SH solve does not reset; see the limitations. The SH appearance coefficients are not included in $G$.

The RGB and $Y$ objectives differ in color basis as well as in channel content, because neither $\mathcal L_1$ nor SSIM is invariant to the YCbCr transform. Two controls bound this confound. A basis control trains the YCbCr SH parameterization under the stock RGB loss, so that only the appearance basis differs from RGB training. A full-YCbCr loss arm, which averages the stock loss over the three transformed channels, changes the scale and structure of the image-space gradients. This changes the densification operating point: how readily the gradients cross the fixed threshold and trigger new Gaussians. We report this arm only as a diagnostic. 

A reduced-luminance control tests whether weaker gradients explain the lower Gaussian count. It multiplies the whole luminance loss, including both $\mathcal L_1$ and D-SSIM, by a fixed per-scene factor $1/s$. Here $s$ is the ratio of the luminance-only to the chroma-only densification statistic, frozen before this arm's outcome was inspected; pilot results from earlier arms were already available. The factors are $1/s=0.411$ on \textsc{counter} and $0.653$ on \textsc{kitchen}. On \textsc{garden} the frozen factor exceeds one, so the arm is not a valid reduced-luminance control there. With the densification threshold unchanged, reducing the luminance loss lowers its densification statistic and produces far fewer Gaussians; the scaling does not establish a match to the chroma-only gradient distribution. A complementary arm multiplies the chroma-only loss by $s$. Both appear in Fig.~\ref{fig:chroma-geometry-controls}, the reduced-luminance control as ``Y-only, down-weighted'' in panel b; details are in the supplement.

\subsection{Appearance solve}

To compare two learned geometries without also comparing the appearance coefficients they happened to learn during training, we re-estimate appearance after freezing $G$. We write $A_S^{(L)}(G)$ for a solve over channel set $S\in\{Y,CbCr,\mathrm{full}\}$ using SH coefficients through degree $L$. Our main operator is
\[
A^* \equiv A_{\mathrm{full}}^{(3)},
\]
which resets and re-estimates all three appearance channels through SH degree 3 while keeping the geometry fixed.

The solve begins with visibility-weighted least squares for each Gaussian. We then render the current solution, measure the remaining image error, and use that residual to update the Gaussians that contributed to those pixels. Because the residual contains every Gaussian's contribution, repeating this step accounts for overlap between Gaussians, which the first per-Gaussian solve ignores. We also use a small regularization term to keep the solution stable. Its strength scales with each Gaussian's total visibility and rises tenfold per SH degree from a base value of $10^{-6}$. Refinement stops when the residual, the coefficient update, or the per-iteration improvement in the residual falls below a relative tolerance (the last for two consecutive iterations), or at an iteration cap. The geometry comparisons use a tolerance of $3\times10^{-4}$ and a 60-iteration cap; the capacity sweep uses $5\times10^{-4}$ and 70 iterations. All 88 solves of the main RGB-versus-$Y$ comparison reach the cap before the tolerance, in both arms alike; the original protocol would have excluded capped solves, and the amendment that retained them is described in the supplement. Running two scenes to 200 iterations changes the RGB--Y gap by at most $0.014$\,dB. The base ridge and the iteration budgets were chosen on \textsc{garden} and \textsc{kitchen} using held-out test-view scores and then fixed for the main geometry comparisons (supplement); the operator ladder and the capacity sweep use the separate settings reported there, and the fit itself uses no test images. Convergence and sensitivity to this regularization are reported in the supplement. Two properties of the solve matter for interpretation. It minimizes squared image error in the YCbCr working space, a quadratic objective close to the RGB squared error that PSNR measures, whereas training minimizes $\mathcal L_1+$D-SSIM; evaluation converts the render to RGB and reports RGB PSNR. It also fits all non-test views, including the eight validation views that training withheld. Every solve uses the same objective and fit views. Comparisons with trained checkpoints also include the effects of this different objective and the additional fit views (Secs.~\ref{sec:luminance-geometry} and~\ref{sec:appearance-capacity}). One $A^*$ solve takes 13--28 minutes on a single NVIDIA L40S (median 19 minutes over the 22 seed-0 base solves), nearly all of it in the refinement; each scene's comparison needs two solves.

\subsection{Metrics}

For scene $k$, we measure the geometry difference as
\begin{equation}
\Delta_{\mathrm{geom},k}
=
Q_k(A^*(G_Y))
-
Q_k(A^*(G_{\mathrm{RGB}})),
\label{eq:geometry-gap-compact}
\end{equation}
where $Q_k$ is held-out RGB PSNR for scene $k$. Before examining the eleven-scene comparison, we set $\pm0.15$\,dB as the minimum threshold for a reconstruction difference large enough to investigate further, and retained it after estimating seed-to-seed variability from the replication. This is an effect-size threshold, not a confidence interval. The comparison was first run at a single seed. Scenes whose seed-0 difference fell outside the band, or within $0.03$ dB of its edge, were replicated with four independent runs per condition, and the remaining scenes were then brought to four runs as well, so that every scene contributes the difference of its condition means. We treat a scene's departure as persistent only if its condition-mean difference remains outside $\pm0.15$\,dB and is larger than twice its standard error.

We also separate the effect of re-estimating appearance from the effect of changing geometry. Let
\[
Q_0=Q(A_{\mathrm{ref}}(G_{\mathrm{RGB}})),\qquad
Q_1=Q(A^*(G_{\mathrm{RGB}})),\qquad
Q_2=Q(A^*(G_Y)),
\]
where $A_{\mathrm{ref}}$ is the appearance stored in the trained RGB checkpoint. The difference $Q_1-Q_0$ shows how much reconstruction quality changes when appearance is re-estimated on the original RGB geometry. The difference $Q_2-Q_1$ then compares RGB- and luminance-trained geometry after both have received the same appearance solve. Scene PSNR is the mean of per-image PSNR over the held-out views, and scene MSE is pooled over the same views, so a mean PSNR change and a mean MSE change do not convert into each other exactly.

We use a similar setup to measure how well appearance can be recovered after individual channels are removed. For channel set $S$, $A_S^{\varnothing}$ denotes the representation after that appearance component has been removed, while $A_{\mathrm{ref}}$ denotes the appearance stored in the trained model. We measure the fraction of the lost reconstruction quality recovered by the solve as
\begin{equation}
R_S^{(L)}
=
\frac{
Q(A_S^{(L)})-Q(A_S^{\varnothing})
}{
Q(A_{\mathrm{ref}})-Q(A_S^{\varnothing})
}.
\label{eq:recovery-compact}
\end{equation}
A value of $R=1$ means that the solve restores the full PSNR difference between the removed state and the trained model. Values above one are possible when re-estimating appearance produces a better reconstruction than the appearance stored in the original checkpoint. Because PSNR is logarithmic, $R$ is a ratio of decibel differences rather than a fraction of recovered squared error, and it can be unstable when its denominator is small. We therefore also report its MSE-domain counterpart,
\[
{\color{black}
R_S^{\mathrm{MSE}}
=
\frac{
\mathrm{MSE}(A_S^{\varnothing})-\mathrm{MSE}(A_S^{(L)})
}{
\mathrm{MSE}(A_S^{\varnothing})-\mathrm{MSE}(A_{\mathrm{ref}})
},
}
\]
for the capacity sweep, together with the per-scene denominators of both ratios for the operator ladder and the sweep, in the supplement.

To measure how much view-dependent appearance is needed, we also compare degree-0 appearance with the full degree-3 representation:
\begin{equation}
H_S
=
Q(A_S^{(3)})-Q(A_S^{(0)}).
\label{eq:higher-order-compact}
\end{equation}
This measures how much reconstruction quality is gained from higher-order SH for luminance or chroma. Its linear counterpart, the relative increase in held-out mean-squared error when degrees 1--3 are withheld, is reported alongside it as a per-scene mean. In these solves the other channel group keeps its trained appearance, so the degree-3 rows of the capacity sweep are not the same solves as $A^*$. On the specular scenes we additionally use a collapse test, which zeroes the higher-order coefficients learned in training and measures the held-out loss without re-solving; unlike the solve, it cannot overfit sparsely observed views.

Before using this measure, we check how much the appearance solver itself can recover. We begin with a single scene-wide color value, which allows only one global correction. We then allow each Gaussian to have its own view-independent DC color. The next solver refines those DC colors together through the rendered image residual, allowing overlapping Gaussians to share the correction. The final solver also includes higher-order SH, allowing appearance to vary with viewing direction. This sequence tells us whether poor recovery comes from the fixed geometry or simply from using an appearance model that is too limited. We separately repeat the solve with SH degrees $L\in\{0,1,2,3\}$ for $Y$ and $CbCr$ to measure how much directional capacity each requires. Both channels use the same geometry and the same camera views, so a difference between them cannot be explained by one channel being observed from more viewing directions than the other.

\subsection{Controls and interventions}

We use additional controls to check whether differences attributed to supervision are instead caused by changes in the number of Gaussians or in the densification process. To control for representation size, if a luminance-trained model contains more Gaussians than its RGB counterpart, we remove the lowest-opacity Gaussians until both models contain the same number and then rerun $A^*$. This opacity-pruning diagnostic replaced the retraining at matched count that was originally planned (supplement). It tests whether the luminance result depends simply on having more primitives. We also compare rendered expected depth between the RGB- and luminance-trained models. Because independently trained 3DGS models vary even under the same condition, we compare these depth differences with the variation observed between repeated runs of the same training condition. The full depth analysis is reported in the supplement.

To test whether luminance itself is more effective than chroma at forming useful geometry, we also train models using chroma alone. We compare $A^*(G_{CbCr})$ with $A^*(G_Y)$ on \textsc{garden}, \textsc{counter}, and \textsc{kitchen}. Chroma-only training produces smaller values of the positional-gradient statistic used by 3DGS to trigger densification. It can therefore produce fewer new Gaussians even if the training schedule is otherwise unchanged. To separate this densification effect from the information carried by chroma, we run an additional chroma condition with a lower densification threshold. The threshold is chosen so that the fraction of Gaussians crossing it at the first densification event matches the luminance condition; all other settings remain unchanged. A densification-free comparison removes the confound outright: both arms start from the same dense point set, taken from a stock RGB run at 15k iterations, and train with densification, pruning, and opacity resets disabled, so that both arms keep the same set of primitives, with the same count and identities, while their positions, shapes, and opacities evolve separately. A contrast-stretched chroma condition scales $Cb$ and $Cr$ to the standard deviation of luminance inside the loss, with the densification threshold re-matched on the stretched statistic, to test whether the D-SSIM term's response to low-contrast chroma explains part of the residual.

Finally, we test whether incorrect chroma can disturb geometry when luminance remains correct. In this condition, luminance is preserved to floating-point precision before quantization, and the recorded uint8 conversion changes it by at most $0.5/255$, while each training view receives another view's chroma through a derangement of the optimization views, in which no view keeps its own chroma and no test or validation view contributes any. Each seed uses its own derangement, generated from its training seed. The luminance structure is therefore unchanged up to rounding, while the chromatic information is deliberately inconsistent across views. This experiment tests whether chroma can substantially alter geometry even when the luminance signal remains available. It is run with four seeds per arm on every scene, against the four-run RGB and luminance-only references of the main comparison. To check the effect of Gaussian count, we prune the lowest-opacity Gaussians from the RGB reference to match the adversarial count whenever the counts differ by more than 5\%.

\section{Luminance-Trained Geometry Supports Nearly the Same Reconstruction}
\label{sec:luminance-geometry}

After the same appearance solve $A^*$, luminance-trained geometry supports nearly the same held-out reconstruction as RGB-trained geometry. At seed 0 (Table~\ref{tab:solver-geometry}), $Q_0$, $Q_1$, and $Q_2$ average $28.112$, $28.555$, and $28.457$ dB, respectively. Re-estimating appearance on the original RGB-trained geometry therefore improves reconstruction by $0.443$ dB on average. Part of this gain is expected from the solve itself: it minimizes squared error in the working space, close to what PSNR measures, whereas training minimizes $\mathcal L_1+$D-SSIM, and it fits the eight validation views that training withheld (Sec.~\ref{sec:experimental-framework}). The gain is mainly in PSNR (SSIM $+0.0022$, LPIPS unchanged). After the same appearance solve is applied to both geometries, the difference over four independent runs per condition on every scene is
\[
{\color{black}
\Delta_{\mathrm{geom}}
=
Q_2-Q_1
=
-0.085\pm0.037~\mathrm{dB},
\qquad
95\%~\mathrm{CI}~[-0.166,\,-0.003],
}
\]
where each scene contributes the difference of its condition means (Table~\ref{tab:four-run-per-scene}) and the standard error is the between-scene standard deviation ($0.121$ dB) divided by $\sqrt{11}$. This is a small deficit of about $0.08$ dB whose 95\% interval just excludes zero. The formal equivalence test at $\pm0.15$ dB narrowly fails: the 90\% interval extends $0.001$ dB beyond the band. We kept the original band. The seed-0 realization gave $-0.098\pm0.042$ dB (Table~\ref{tab:solver-geometry}; supplement).
In SSIM the seed-0 comparison gives $-0.0011$, and in LPIPS $-0.0011$, where lower LPIPS favors luminance-trained geometry (Table~\ref{tab:solver-geometry}; per scene in Table~\ref{tab:solver-geometry-ssim-lpips}). SSIM therefore moves in the same direction as PSNR, on seven of eleven scenes, while LPIPS favors luminance-trained geometry on average and RGB-trained geometry on \textsc{playroom}, \textsc{dr.~johnson}, and \textsc{train}. Neither metric was replicated across seeds. Comparing $Q_2$ directly with the original RGB-trained checkpoint would mix the effect of changing geometry with the improvement from re-estimating appearance. The same operator recovers omitted chroma on fixed geometry almost completely (Sec.~\ref{sec:appearance-capacity}), which is why we trust it as the common reference. Both geometries use the same fit views, objective, regularization, and stopping rule. All 22 base solves reach the 60-iteration cap before satisfying the stopping tolerance. Running the solve to 200 iterations with stopping disabled on \textsc{kitchen} and \textsc{counter} moves the gap by only $-0.014$ and $-0.010$ dB, because both arms keep improving together. This supports stability of the gap on these two scenes, but does not establish convergence on every scene.

\begin{table}[t]
\centering
\small
\caption{\textbf{Reconstruction before and after the common appearance solve.}
Means are over eleven seed-0 scenes; this is the single-seed realization. The four-run results supporting the headline are in Table~\ref{tab:four-run-per-scene}. The same $A^*$ is applied to the last three rows. The basis control trains the YCbCr SH parameterization under the stock RGB loss, so that only the appearance basis differs from RGB training. SSIM and LPIPS are computed on the same held-out views and render path as PSNR.}
\label{tab:solver-geometry}
\begin{tabular}{@{}lrrrr@{}}
\toprule
& \multicolumn{2}{c}{\textbf{PSNR}} & & \\
\cmidrule(lr){2-3}
\textbf{Geometry} & dB & vs.\ $Q_1$ & \textbf{SSIM}\,$\uparrow$ & \textbf{LPIPS}\,$\downarrow$\\
\midrule
\multicolumn{5}{@{}l}{\emph{Trained appearance}}\\
$G_{\mathrm{RGB}}$ & 28.112 & -- & 0.8694 & 0.1469\\
\addlinespace
\multicolumn{5}{@{}l}{\emph{Common appearance solve $A^*$}}\\
$G_{\mathrm{RGB}}$ & 28.555 & 0.000 & 0.8717 & 0.1469\\
$G_Y$ & 28.457 & $-0.098$ & 0.8706 & 0.1458\\
\addlinespace
\multicolumn{5}{@{}l}{\emph{Basis control, same $A^*$}}\\
$G_{\mathrm{YCbCr}}$ & 28.537 & $-0.018$ & 0.8719 & 0.1464\\
\bottomrule
\end{tabular}
\end{table}

A basis control separates channel content from color basis. Training the YCbCr SH parameterization under the stock RGB loss gives a matched-appearance difference of $-0.018\pm0.041$ dB relative to RGB training (95\% CI $[-0.109,+0.073]$), with Gaussian counts within $0.967$--$1.020$ of the RGB count on all eleven scenes. The prespecified rule for this control had a mean clause and a count clause. The mean clause is met: the difference lies inside the band, so the appearance basis does not account for the geometry difference in the mean. The count clause compared single-seed counts: at a single seed the control places eight scenes inside the band against nine for the seed-0 main comparison, because \textsc{kitchen} moves in while \textsc{train} and \textsc{playroom} move out. The rule defined no outcome for this combination, so we report both results without a verdict. The count is secondary in any case, because the main comparison is summarized by its four-run mean.

The full-YCbCr loss arm behaves differently: it densifies to $14\%$--$40\%$ of the RGB count on every scene, and on \textsc{counter} its reconstruction ($29.121$ dB) matches the reduced-luminance control of Sec.~\ref{sec:experimental-framework} ($29.037$ dB) within $0.15$ dB, consistent with a deficit caused partly by fewer Gaussians. Averaging the stock loss over the three transformed channels changes both the scale and the structure of the image-space gradients, so the fixed densification threshold no longer represents the same operating point. The arm therefore measures how the densifier responds to a change of loss basis, not a limit of the YCbCr parameterization, which the basis control shows supports nearly the same reconstruction as RGB.

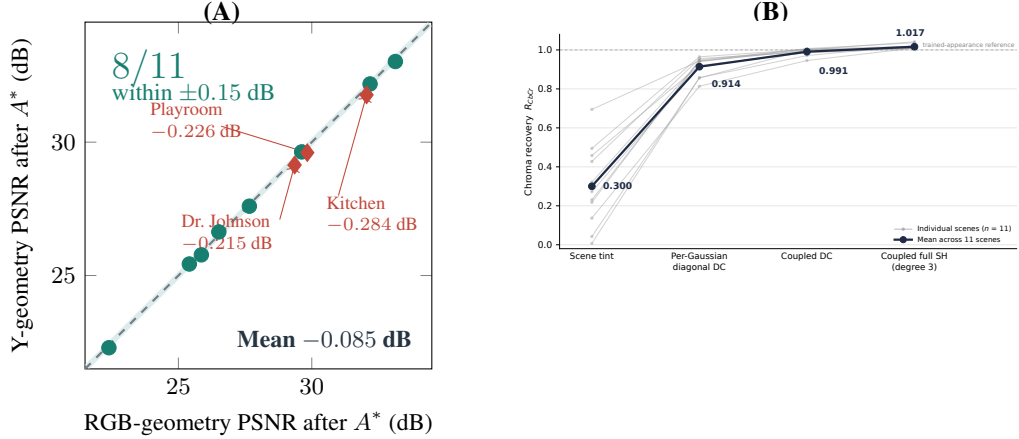
\begin{figure}[h!]
\centering

\begin{minipage}[t]{0.48\textwidth}
\centering
\textbf{(A)}\\[-0.5ex]

\begin{tikzpicture}
\begin{axis}[
    width=\linewidth,
    height=0.92\linewidth,
    xmin=21.5,
    xmax=34.5,
    ymin=21.5,
    ymax=34.5,
    axis equal image,
    xlabel={RGB-geometry PSNR after $A^*$ (dB)},
    ylabel={Y-geometry PSNR after $A^*$ (dB)},
    tick label style={font=\small},
    label style={font=\small},
    clip=true
]

\addplot[
    draw=none,
    fill=solvergreen!15
]
coordinates {
    (21.5,21.65)
    (34.35,34.5)
    (34.5,34.35)
    (21.65,21.5)
} -- cycle;

\addplot[
    dashed,
    thick,
    darktext!65,
    domain=21.5:34.5
]
{x};

\addplot[
    only marks,
    mark=*,
    mark size=2.7pt,
    solvergreen
]
coordinates {
    (25.409,25.432)
    (33.133,33.023)
    (29.621,29.634)
    (27.660,27.600)
    (32.185,32.182)
    (26.515,26.635)
    (22.395,22.293)
    (25.860,25.775)
};

\addplot[
    only marks,
    mark=diamond*,
    mark size=3.3pt,
    warningred
]
coordinates {
    (32.056,31.772)
    (29.359,29.144)
    (29.832,29.606)
};

\node[
    anchor=north west,
    font=\Large\bfseries,
    text=solvergreen
]
at (rel axis cs:0.05,0.95)
{$8/11$};

\node[
    anchor=north west,
    font=\small,
    text=solvergreen
]
at (rel axis cs:0.05,0.85)
{within $\pm0.15$ dB};

\node[
    font=\scriptsize,
    text=warningred,
    align=left,
    anchor=west
]
at (axis cs:30.2,27.3)
{Kitchen\\$-0.284$ dB};

\draw[
    warningred,
    ->,
    thin
]
(axis cs:31.0,28.0)
--
(axis cs:32.056,31.772);

\node[
    font=\scriptsize,
    text=warningred,
    align=left,
    anchor=east
]
at (axis cs:28.9,26.6)
{Dr.~Johnson\\$-0.215$ dB};

\draw[
    warningred,
    ->,
    thin
]
(axis cs:28.8,27.1)
--
(axis cs:29.359,29.144);

\node[
    font=\scriptsize,
    text=warningred,
    align=left,
    anchor=west
]
at (axis cs:23.6,30.8)
{Playroom\\$-0.226$ dB};

\draw[
    warningred,
    ->,
    thin
]
(axis cs:26.6,30.6)
--
(axis cs:29.832,29.606);

\node[
    anchor=south east,
    font=\small\bfseries,
    text=darktext
]
at (rel axis cs:0.97,0.03)
{Mean $-0.085$ dB};

\end{axis}
\end{tikzpicture}
\end{minipage}
\hfill
\begin{minipage}[t]{0.48\textwidth}
\centering
\textbf{(B)}\\[-0.5ex]
\includegraphics[width=\linewidth]
    {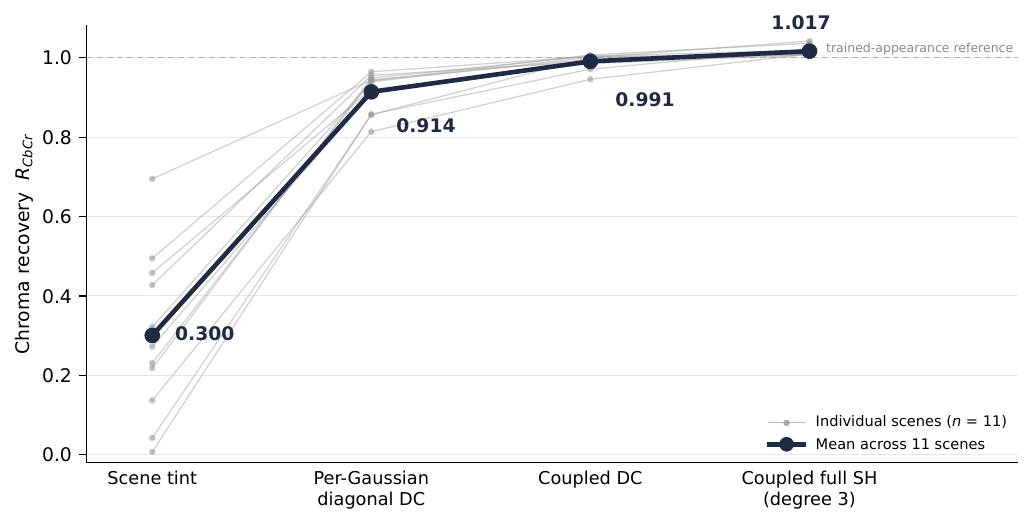}
\end{minipage}

\caption{
\textbf{Luminance-trained geometry approaches RGB-trained geometry under the same appearance solve, and chroma can be recovered once geometry is fixed.}
\textbf{(A)} Each point compares held-out RGB reconstruction after applying the same full-appearance solve $A^*$ to RGB-trained and luminance-trained geometry. The dashed line marks exact parity, and the shaded region is the prespecified $\pm0.15$ dB investigation band. Every point is the condition mean over four independent runs per condition, so the mean difference of the points is the eleven-scene mean; the labels give the condition-mean differences for the three departures. Eight of eleven scenes lie inside the band; \textsc{kitchen}, \textsc{dr.~johnson}, and \textsc{playroom} retain RGB-geometry advantages.
\textbf{(B)} Chroma recovery from fixed RGB-trained geometry improves as the appearance solve moves from a single scene-wide tint to independent per-Gaussian DC, coupled DC, and coupled degree-3 SH. The dashed line at $R_{CbCr}=1$ marks the reconstruction quality of the appearance stored in the trained model.
}
\label{fig:geometry-and-recovery}
\end{figure}

\begin{figure}[h!]
    \centering

    \begin{subfigure}[t]{0.98\textwidth}
        \centering
        \includegraphics[width=\linewidth]
        {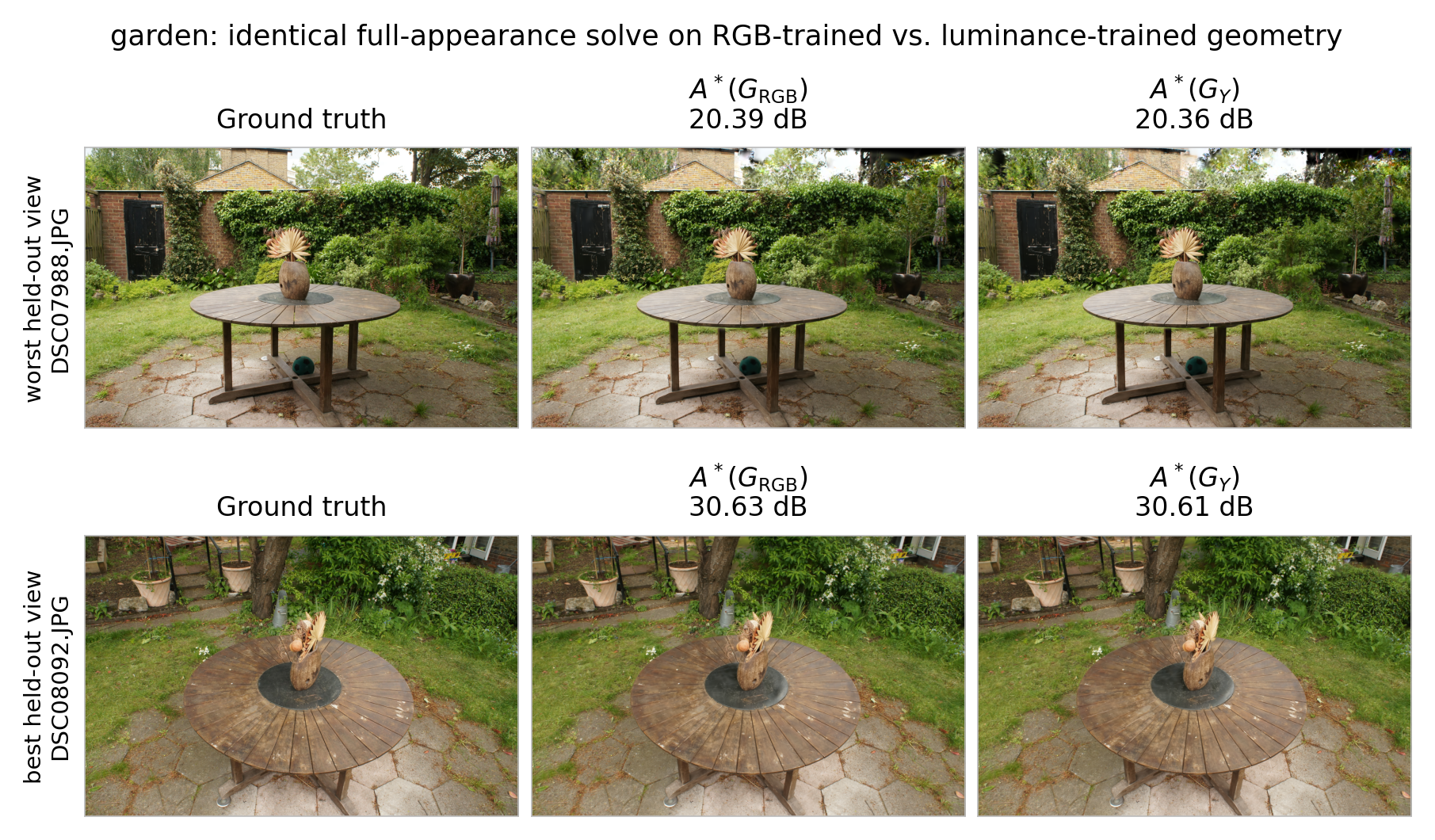}
        \caption{\textsc{garden}, an in-band scene.}
    \end{subfigure}

    \vspace{0.8em}

    \begin{subfigure}[t]{0.98\textwidth}
        \centering
        \includegraphics[width=\linewidth]
        {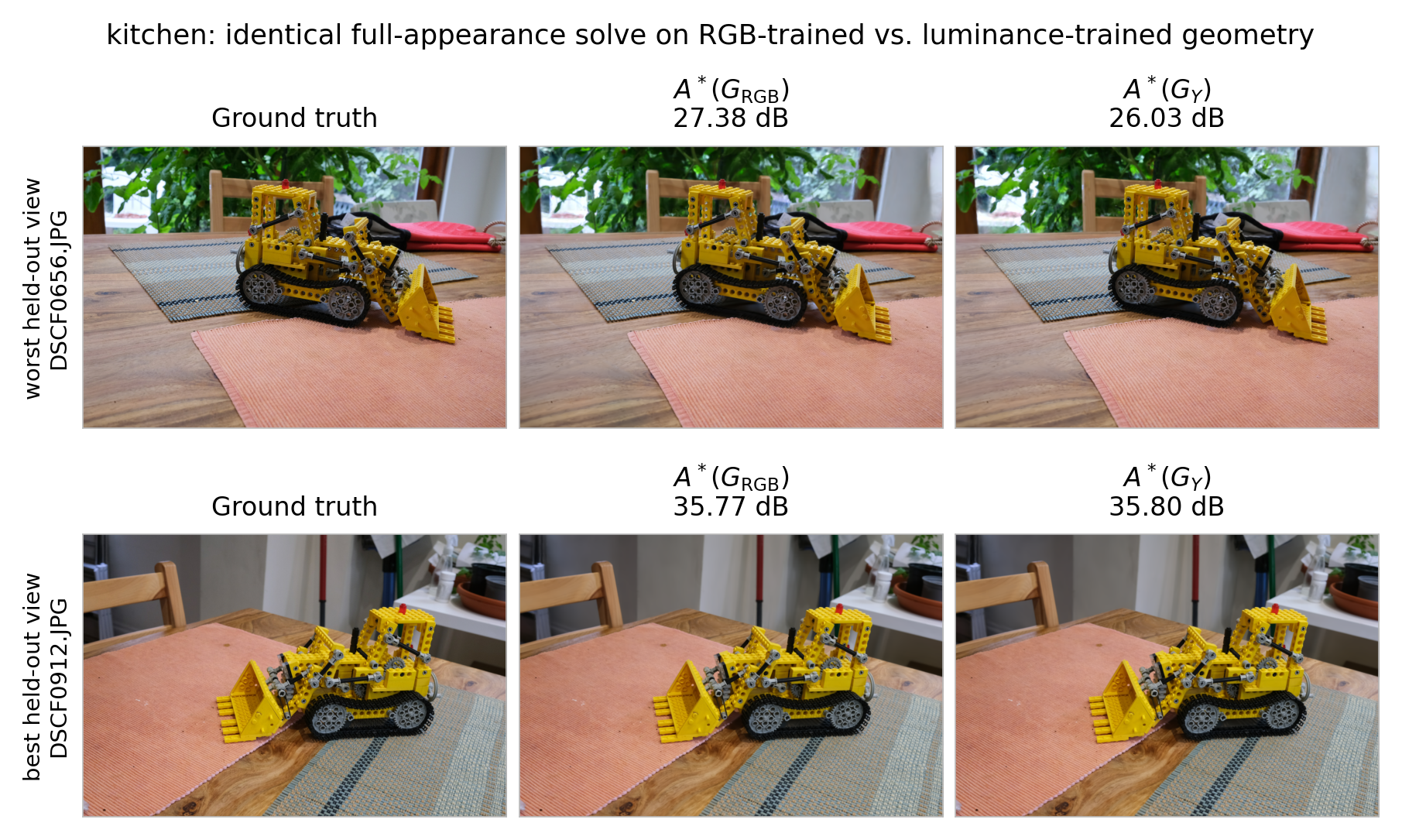}
        \caption{\textsc{kitchen}, an outside-band scene.}
    \end{subfigure}

    \caption{
    \textbf{Matched appearance fitting exposes both the typical
    near-equivalence and a scene-specific departure.}
    Each row compares ground truth with the identical canonical
    full-appearance solve \(A^*\) applied to RGB-trained geometry
    \(G_{\mathrm{RGB}}\) and luminance-trained geometry \(G_Y\).
    \textbf{(a)} On \textsc{garden}, the two geometries support nearly
    identical held-out reconstructions on both the weakest and strongest views
    shown: \(20.39\) versus \(20.36~\mathrm{dB}\), and \(30.63\) versus
    \(30.61~\mathrm{dB}\), respectively.
    \textbf{(b)} \textsc{kitchen} illustrates why benchmark-wide
    near-equivalence is not universal. Its weakest shown view favors
    RGB-trained geometry by \(1.35~\mathrm{dB}\)
    (\(27.38\) versus \(26.03~\mathrm{dB}\)), while the strongest shown view
    is effectively tied (\(35.77\) versus \(35.80~\mathrm{dB}\)).
    All reconstructions use the same appearance-fitting procedure; only the
    frozen geometry differs.
    }
    \label{fig:qualitative-common-fit}
\end{figure}

With four runs per condition on every scene (Fig.~\ref{fig:geometry-and-recovery}A), eight of the eleven scenes lie inside the band. Their condition-mean differences range from $-0.110$ dB on \textsc{bonsai} to $+0.119$ dB on \textsc{stump} and average $-0.026$ dB, so the overall mean is not produced by cancellation between large opposite effects. Three scenes lie outside the band. \textsc{kitchen} is at $-0.284\pm0.103$ dB; its effect is heterogeneous across runs, with same-seed-label differences spanning $-0.452$ to $+0.046$ dB, so we read it as a persistent average advantage rather than one reproduced in every run. \textsc{dr.~johnson} is at $-0.215\pm0.041$ dB and \textsc{playroom} at $-0.226\pm0.054$ dB. Among the in-band scenes, \textsc{bonsai} ($-0.110\pm0.035$ dB) and \textsc{train} ($-0.102\pm0.051$ dB) lie nearest the lower edge, \textsc{counter} is at parity ($+0.013\pm0.014$ dB), and \textsc{stump} favors luminance-trained geometry. The seed-0 realization had placed nine scenes in the band; replication moved \textsc{playroom} out and changed no other status. Replicating the scenes nearest the band edge first and the remaining scenes afterward shifted the mean by $0.009$ dB, so the order of replication did not shape the result.
We therefore describe luminance-trained geometry as supporting nearly the same reconstruction as RGB-trained geometry, with a small average deficit and three replicated scene-specific departures of $0.2$--$0.3$ dB, rather than as equivalent to it. Figure~\ref{fig:qualitative-common-fit} shows both sides of the result on held-out views.

We use two additional checks to make sure that the small reconstruction difference is not caused by other changes in the representation. At seed 0, four scenes contain more Gaussians after luminance-only training than after RGB training. When the lowest-opacity Gaussians are removed until the primitive counts match and $A^*$ is run again, held-out PSNR changes by only $0.000$--$0.005$ dB. The RGB-versus-$Y$ difference changes by at most $0.006$ dB. The result therefore does not depend on luminance-trained models having more Gaussians.

We also compare rendered expected depth between independently trained models. On the replicated in-band scenes, the depth differences between RGB- and luminance-trained models are similar to the differences between repeated runs of the same condition. \textsc{kitchen} and \textsc{dr.~johnson}, the two scenes replicated in this analysis, do not share the same depth discrepancy, and depth difference does not track the reconstruction difference across the benchmark ($\rho_{\mathrm{Spearman}}=0.064$, $n=11$). These results support the narrower conclusion measured by $A^*$: luminance- and RGB-trained geometries usually support very similar held-out reconstruction, even though the individual Gaussian parameters need not be identical.

\section{Chroma Is Recoverable on Fixed Geometry and Uses Little Higher-Order Capacity}
\label{sec:appearance-capacity}

We next ask how much chromatic appearance can be recovered once geometry has been fixed. Starting from RGB-trained models, we remove the $CbCr$ appearance while retaining geometry and luminance, then re-estimate chroma from the training images. Mean recovery across the eleven benchmark scenes increases from
\[
0.300\rightarrow0.914\rightarrow0.991\rightarrow1.017
\]
as more appearance capacity is added to the solve (Fig.~\ref{fig:geometry-and-recovery}B). The denominators of these ratios are not small: removing chroma costs between $3.1$ dB (\textsc{bicycle}) and $16.5$ dB (\textsc{playroom}) of held-out PSNR, $8.2$ dB on average, so no scene's ratio rests on a small difference; per-scene values and the MSE-domain denominators are in the supplement (Table~\ref{tab:recovery-ladder}).

The first solve assigns a single chromatic value to the entire scene. The second gives each Gaussian its own view-independent DC chroma but estimates each Gaussian independently. Coupled DC uses the same degree-0 appearance representation but updates the Gaussians together through the rendered image residual, accounting for overlap between their contributions. The final solve adds SH degrees 1--3 and therefore allows chromatic appearance to change with viewing direction. Recovery increases at every step on every scene.

The large improvement from a scene-wide tint to per-Gaussian DC shows that chroma varies spatially across the scene. The improvement from independent to coupled DC shows that overlapping Gaussians need to be considered together when appearance is re-estimated. Adding higher-order SH closes the remaining difference. Values slightly above $R_{CbCr}=1$ mean that the new appearance solve reconstructs the held-out images slightly better than the appearance stored in the original trained model, which is possible because the solve minimizes squared error directly and fits more views than training used.

This experiment does not predict chroma from geometry. Every solve has access to the target chromatic training images. Recovery near one means that the fixed geometry can support the removed chromatic appearance once that appearance is re-estimated. It does not mean that chroma is stored in luminance or can be inferred from geometry alone. The same result holds across additional checkpoints and independently trained reference models; those results are reported in the supplement.

We then measure how much higher-order SH contributes to luminance and chroma. Using the notation defined in Sec.~\ref{sec:experimental-framework}, the average contributions across the eleven benchmark scenes are
\[
H_Y=1.441~\mathrm{dB},
\qquad
H_{CbCr}=0.187~\mathrm{dB}.
\]
The ordering $H_Y>H_{CbCr}$ holds on every benchmark scene. In linear terms, removing SH degrees 1--3 raises held-out MSE by $35.0\%$ for luminance but by $3.6\%$ for chroma, as arithmetic means over scenes (medians $23.9\%$ and $1.9\%$); these are means of pooled per-scene MSE ratios, whereas the decibel values are means of per-image PSNR, so the two do not convert into each other exactly. The luminance values are skewed, from $1.7\%$ on \textsc{playroom} to $97.5\%$ on \textsc{bonsai}. The asymmetry holds in every metric: degrees 1--3 add $+0.028$ SSIM and improve LPIPS by $0.017$ for luminance, against $+0.004$ and $0.004$ for chroma (Table~\ref{tab:capacity-ssim-lpips}). The same difference appears in the recovery ratios. With degree-0 appearance alone, chroma already reaches $R_{CbCr}^{(0)}=0.994$, while luminance reaches $R_Y^{(0)}=0.924$.\footnote{The operator ladder gives $0.991$ for coupled DC because it uses earlier ridge and tolerance settings; see the supplement.} By degree 3, both are approximately $1.017$. The MSE-domain ratio of Sec.~\ref{sec:experimental-framework} gives the same ordering: $0.990$ for luminance and $1.000$ for chroma at degree 0, and $1.002$ and $1.007$ at degree 3 (Table~\ref{tab:recovery-mse}). Both groups reach approximately the trained-reference quality at degree 3 under this solver, so neither group is left with a recovery deficit. Most of the chromatic reconstruction quality is therefore available from view-independent per-Gaussian color, while luminance gains substantially more from higher-order directional appearance.

\begin{figure}[h!]
    \centering
    \includegraphics[width=\textwidth]
        {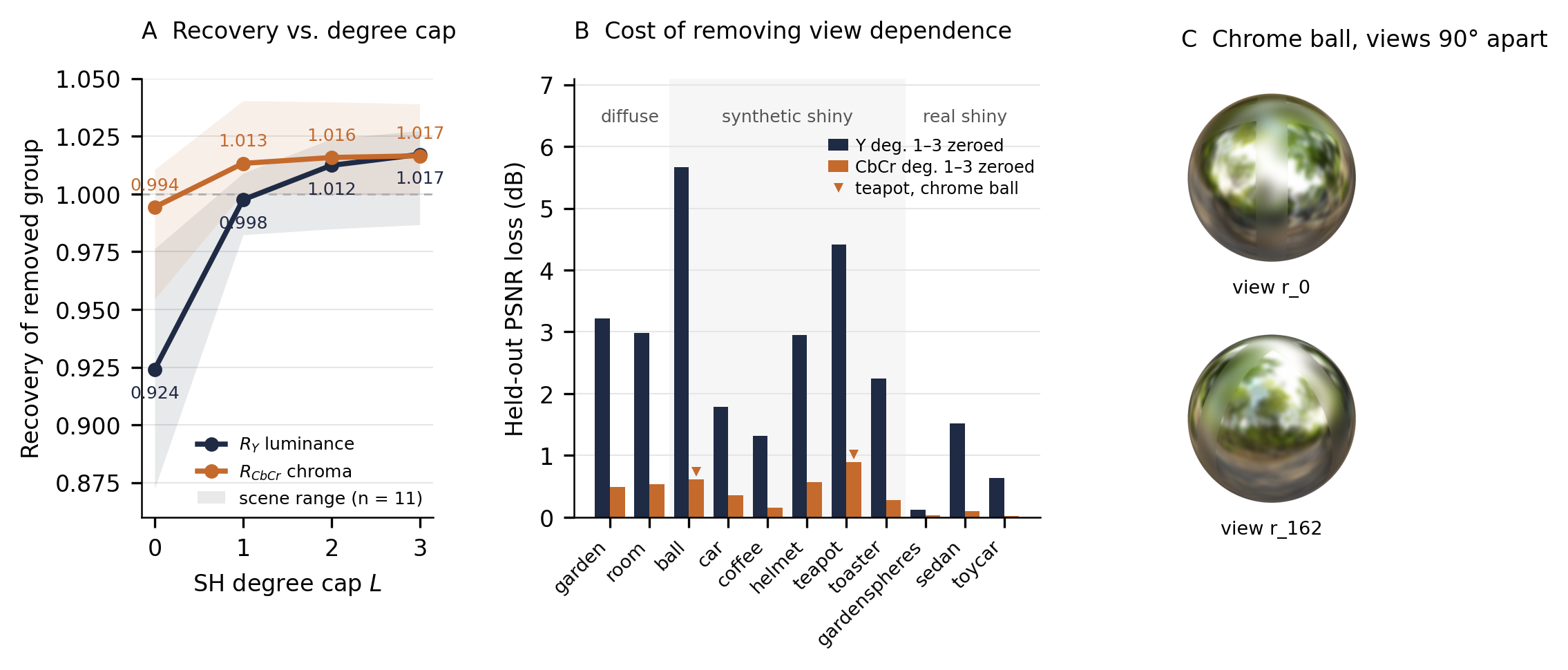}
    \caption{
    \textbf{Higher-order SH contributes far more held-out reconstruction quality to luminance than to chroma.}
    \textbf{(A)} Recovery as the SH degree cap increases across the eleven benchmark scenes, measured by re-solving at each cap. Chroma is already close to its full recovery at degree 0, while luminance gains substantially more from higher-order SH.
    \textbf{(B)} Held-out PSNR loss after removing SH degrees 1--3 from luminance or chroma on diffuse, synthetic-shiny, and real-shiny scenes, measured by the collapse test of Sec.~\ref{sec:experimental-framework}, which zeroes the trained coefficients without re-solving. Higher-order luminance is more important in every case, although shiny scenes also show measurable chromatic losses.
    \textbf{(C)} Two ground-truth views of the chrome ball separated by approximately $90^\circ$. Reflected colors move substantially across the surface with viewpoint, motivating the fixed-surface-point chromaticity analysis.
    }
    \label{fig:directional-capacity}
\end{figure}

Strongly specular scenes show the limit of this pattern. Higher-order chroma still contributes less reconstruction quality than higher-order luminance, but its contribution is not negligible. Removing higher-order chroma costs about $0.9$ dB on \textsc{teapot} and $0.61$ dB on the chrome ball, against about $4.4$ and $5.7$ dB for luminance. The ordering also holds when each channel group's higher-order gain is divided by its total recoverable PSNR gain. This compares the higher-order benefit relative to what each group can recover. SH degrees 1--3 account for $0.116$ of luminance's recoverable span over nine specular scenes, against $0.034$ for chroma over eight; \textsc{gardenspheres} chroma is excluded because its recoverable span is below $1$ dB, which makes the ratio unstable. On the same eight scenes the luminance share is $0.134$. The chrome ball gives $0.428$ against $0.125$.

Two real captures, \textsc{gardenspheres} and \textsc{toycar}, disagree in sign with the collapse test (Sec.~\ref{sec:experimental-framework}), which zeroes the trained higher-order coefficients. On both, the degree-0 luminance solve already exceeds the trained model ($R_Y^{(0)}=1.079$ and $1.030$), and adding degrees 1--3 to the solve lowers held-out PSNR by $0.152$ and $0.046$ dB. This is consistent with overfitting of higher-order coefficients when the least-squares solve has few training views. The collapse test avoids this additional fitting by measuring the coefficients learned during training. The two tests agree in sign on the other seven scenes. 

A direct measurement on the chrome ball shows how much color at a fixed surface point changes across views, relative to how much color varies across the surface in one view. At fixed surface points, the across-view standard deviations of $Cb$ and $Cr$ ($0.019$ and $0.011$) are $57\%$ and $62\%$ of their spatial spread within a view, compared with $47\%$ for $Y$ (across-view $0.103$). Chroma therefore varies substantially with viewpoint relative to its own spatial spread. Removing higher-order chroma has a smaller reconstruction cost because the chromatic channels have lower amplitude and view-independent per-Gaussian color captures most of their recoverable quality.

\section{Geometry Formation Is Luminance-Dominated}
\label{sec:channel-geometry}

The matched-appearance comparison shows that luminance-only training often forms enough support for reconstruction, but it does not establish whether chroma could do the same job. We therefore train matched $Y$-only and $CbCr$-only models and evaluate both through $A^*$. The RGB and luminance-only reference arms in this section were retrained with the logging needed for the densification analysis; their $A^*$ values differ from the corresponding Table~\ref{tab:per-scene-q012} rows by at most $0.129$ dB, and the Table~\ref{tab:per-scene-q012} rows are unchanged. On \textsc{garden}, \textsc{counter}, and \textsc{kitchen}, chroma-only geometry is $4.035$, $3.865$, and $5.540$ dB below luminance-only geometry, respectively. Because appearance is re-estimated identically, this multi-decibel gap reflects the representation formed during training rather than the chromatic appearance stored by the model.

\begin{figure}[h!]
    \centering

    \begin{minipage}[t]{0.48\textwidth}
        \centering
        \includegraphics[width=\linewidth]
            {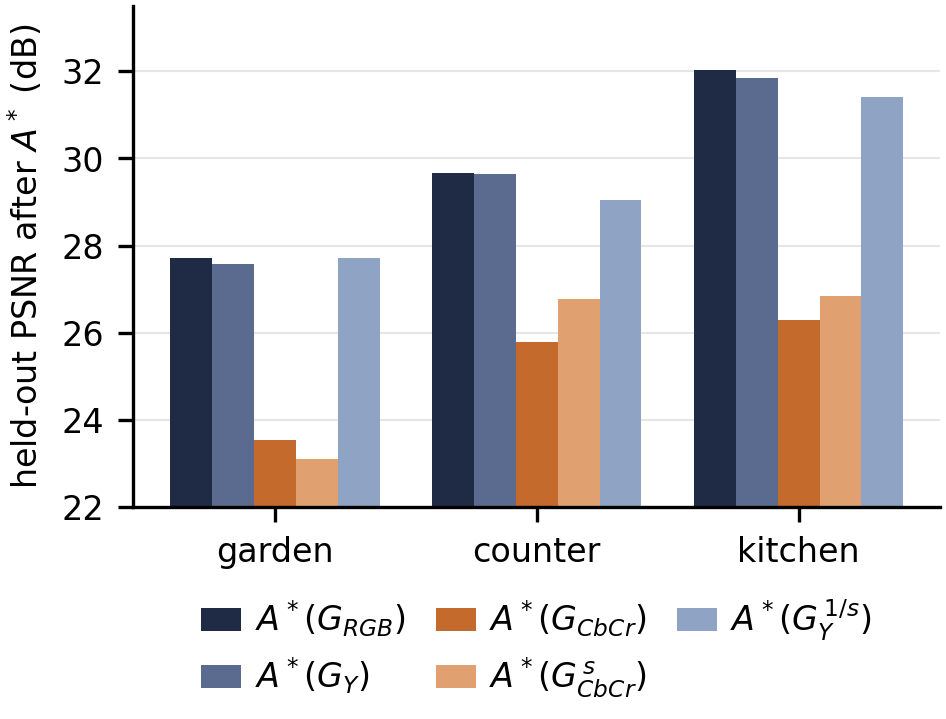}
        \textbf{(a)}
    \end{minipage}
    \hfill
    \begin{minipage}[t]{0.48\textwidth}
        \centering
        \includegraphics[width=\linewidth]
            {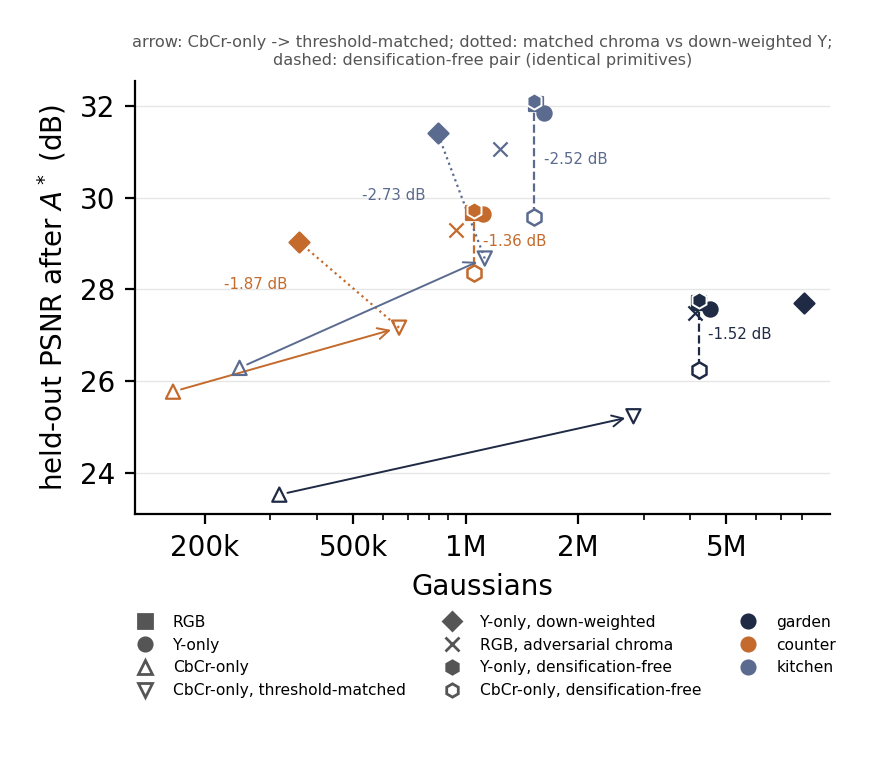}
        \textbf{(b)}
    \end{minipage}

    \caption{
    \textbf{Chromatic supervision does not substitute for luminance when
    learning reconstruction-supporting geometry.}
    \textbf{(a)} Held-out reconstruction after the identical canonical
    full-appearance solve \(A^*\). Luminance-trained geometry remains close
    to RGB-trained geometry, whereas raw chroma-only geometry loses
    \(3.9\)--\(5.5~\mathrm{dB}\). The five bars per scene are, in legend order, $A^*(G_{\mathrm{RGB}})$ and $A^*(G_Y)$ from the reruns of this section, raw chroma-only $A^*(G_{CbCr})$, chroma-only with the loss multiplied by $s$ ($A^*(G^{s}_{CbCr})$), and luminance-only with the loss multiplied by $1/s$ ($A^*(G^{1/s}_{Y})$), the reduced-luminance control of Sec.~\ref{sec:experimental-framework}; on \textsc{garden} the two loss-scaled arms are not valid controls (supplement).
    \textbf{(b)} Reconstruction quality versus Gaussian count for the
    channel interventions and capacity controls, including the
    threshold-matched and densification-free chroma conditions. Lowering chroma's densification
    threshold increases both primitive count and reconstruction quality but
    does not close the gap to luminance. On \textsc{counter} and
    \textsc{kitchen}, luminance-supervised models remain substantially better
    even when they contain fewer primitives. Dashed lines join the
    densification-free pairs, which keep the same set of primitives; chroma-only
    geometry still trails by \(1.4\)--\(2.5~\mathrm{dB}\). ``Y-only, down-weighted'' is the reduced-luminance control of Sec.~\ref{sec:experimental-framework}.
    }
    \label{fig:chroma-geometry-controls}
\end{figure}

Raw chroma-only training also exposes an optimization confound: under the stock absolute densification threshold, the chroma arms finish with only $7.0\%$--$15.3\%$ as many Gaussians as their $Y$ counterparts. Lowering the chroma threshold to match the initial threshold-crossing fraction moves chroma into a comparable growth regime and improves reconstruction substantially. Yet the corrected models remain $2.330$, $2.472$, and $3.160$ dB below $Y$ on the three scenes, closing only $36\%$--$43\%$ of the original gap. Scaling the chroma-only loss by $s$ instead of lowering the threshold recovers less, $+1.005$ dB on \textsc{counter} and $+0.539$ dB on \textsc{kitchen} over raw chroma-only; on \textsc{garden} the frozen factor fell below one, so that arm down-weights chroma and is not a valid control there (Fig.~\ref{fig:chroma-geometry-controls}a). Removing the densifier entirely gives the same answer. Starting both arms from the dense point set of a stock RGB run at 15k iterations and training for 30k iterations with densification, pruning, and opacity resets disabled, so that both arms keep the same set of $4{,}218{,}069$, $1{,}050{,}890$, and $1{,}530{,}247$ primitives (same count and identities, separately evolving parameters), chroma-only geometry remains $1.519$, $1.363$, and $2.519$ dB below luminance-only geometry after $A^*$. The dense initialization helps the chroma arm far more than the luminance arm ($+0.91$ to $+1.20$ dB against $+0.09$ to $+0.27$ dB), consistent with too few Gaussians explaining part, but not all, of the gap. The stock densifier therefore amplifies the asymmetry but does not create it. The step from the threshold-matched to the densification-free comparison also changes the initialization, so its further $0.6$--$1.1$ dB cannot be assigned to the densifier alone. Because the dense set was placed under RGB supervision, which includes luminance, the chroma arm starts from support it did not form itself. We therefore read the remaining deficit as holding under this shared RGB-placed initialization, rather than as a bound on chroma-only training in general (Fig.~\ref{fig:chroma-geometry-controls}b).

Chroma nevertheless participates in geometry formation. In the adversarial condition, luminance is preserved to within uint8 rounding while each view receives spatially and multi-view inconsistent chroma from another optimization view, under a derangement drawn afresh for each seed. Across four seeds per arm on every scene, adversarial chroma lowers common-$A^*$ reconstruction relative to the luminance-only reference by $0.145\pm0.020$, $0.248\pm0.037$, and $0.665\pm0.097$ dB on \textsc{garden}, \textsc{counter}, and \textsc{kitchen}, with complete separation: every adversarial seed scores below every reference seed. Because each seed carries its own mapping, these standard errors include the variation between mappings.
The smaller Gaussian count does not explain this effect in the tested pruning control. The adversarial arms end with fewer Gaussians, but pruning the RGB reference to the adversarial count costs only $0.002$ dB on \textsc{counter} and $0.016$ dB on \textsc{kitchen}. These perturbations are much smaller than the $3.9$--$5.5$ dB loss from removing luminance, but they reject the stronger claim that chroma is geometrically inert. Under the tested 3DGS objective, geometry formation is therefore luminance-dominated rather than luminance-exclusive.

We next test whether low chromatic contrast explains the remaining deficit. Two parts of the training loop respond to contrast differently. To the extent the Adam optimizer normalizes gradient scale, rescaling a channel by a constant would change its update steps very little. The densifier does not normalize: it thresholds the raw accumulated positional gradient, which is why low chroma contrast acts mainly through densification and why the threshold-matched arm recovers as much as it does. What remains scale-dependent in the loss is the D-SSIM term, whose constants are fixed in image units, together with the relative weighting of $Cb$ and $Cr$. The contrast-stretched arm tests that remaining part. It does not change the information content or the signal-to-noise ratio of the chromatic channels, so it cannot tell us whether the relevant variable is luminance itself or spatial contrast. A chroma-only arm with $Cb$ and $Cr$ stretched to the standard deviation of luminance inside the loss, with the densification threshold re-matched on the stretched statistic, recovers $+0.206$, $+0.072$, and $+0.091$ dB on the three scenes (mean $+0.123$ dB), at most $9\%$ of the residual; $2.1$--$3.1$ dB remain. Equalizing chromatic contrast therefore accounts for a small, nonzero part of the deficit, and the remainder is a property of the chromatic signal as captured in these datasets, under the tested loss and optimizer (Sec.~\ref{sec:discussion} discusses the capture pipeline). 

A synthetic scene tests a boundary visible only through chroma. It contains a plate in front of a backdrop: the upper rim has only a chroma step, while the lower rim has both chroma and luminance steps. The prespecified criterion counts a pixel as failed when its estimated depth is nearer the wrong surface. The difference-in-differences between the isoluminant and luminance rims is $+0.028$, just below the $0.03$ cutoff, so this criterion does not confirm failure. Post hoc, luminance-only training pulls depth $19\%$--$30\%$ toward the backdrop across the isoluminant rim against $4.5\%$ across the luminance rim, whereas RGB and chroma-only training stay at $2\%$; the boundary is blurred rather than misplaced.

\section{Delayed Chroma Is Consistent with the Measured Asymmetry}
\label{sec:training-dynamics}

The delayed-chroma observation that motivated the study becomes easier to interpret after the geometry and appearance tests. On replicated runs, withholding chroma for the first half of training largely preserves final quality on \textsc{garden} ($+0.084$ dB, SD $0.081$) and \textsc{stump} ($+0.029$ dB, SD $0.148$), while \textsc{bonsai} retains a $-0.268$ dB deficit (SD $0.073$). These SDs describe the paired differences over three runs per arm. Thus the robust result is preservation on some scenes, not a general improvement over RGB training.

When chroma is released, held-out PSNR rises by about $1.8$ dB over the next 2k iterations. Moving release from 10k to 20k moves the jump with it, whereas 2{,}000-iteration windows around the end of densification show changes of only $0.10$--$0.13$ dB. An independent FastGS implementation reproduces the pattern at a different scale (release step $1.255$ dB, SD $0.057$ over its runs), and repeating Delayed Chroma in BT.601, BT.709, YCoCg, and YIQ meets both the prespecified final-quality and release-jump criteria in 36 of 44 scene--color-space combinations. The release jump is expected because previously unsupervised channels begin to improve, so it is not itself evidence for geometry equivalence. Chroma can be fitted rapidly even while geometry and luminance appearance continue to change. Freezing both at the release and training only chromatic appearance to 30k costs $0.809$ dB on average over the eleven scenes when only the chromatic DC term trains, and $0.602$ dB when all chromatic SH train (from $+0.12$ on \textsc{stump} to $-1.72$ dB on \textsc{room}), showing that continued updates to geometry and luminance appearance together improve reconstruction after release.

Continuing one luminance-only checkpoint from 15k to 30k twice, once without and once with chromatic supervision, with the same set of primitives in both branches because densification has ended, changes $A^*$-supported reconstruction by $+0.113$, $+0.060$, and $+0.118$ dB (mean $+0.097$ dB), and the three-scene mean of the chroma branch is $0.001$ dB above the three-scene mean of the four-run RGB arm (Table~\ref{tab:four-run-per-scene}), with per-scene differences of $+0.014$, $+0.077$, and $-0.087$ dB. In these forks, adding chroma improves reconstruction after the common appearance solve by about $0.1$ dB, close to the RGB advantage of Sec.~\ref{sec:luminance-geometry}. All three changes are positive, but each comes from a single checkpoint, so these estimates are suggestive. Full boundary controls and release-time sweeps are in the supplement.

\section{Discussion}
\label{sec:discussion}

The experiments identify two distinct luminance--chroma asymmetries in trained 3DGS. Along the geometry axis, luminance-only supervision usually forms a non-appearance representation that supports nearly the same reconstruction as RGB after SH appearance is re-estimated identically, whereas chroma-only supervision forms a substantially weaker representation. Along the appearance axis, higher-order SH contributes far more reconstruction quality to luminance than to chroma. These axes are related but should not be collapsed into the claim that luminance is ``geometry'' and chroma is ``color.'' RGB retains reproducible scene-specific geometry advantages, conflicting chroma can perturb optimization, and view-dependent chromatic reflection can require higher-order SH. With four runs per condition on every scene, the eleven-scene gap is $-0.085\pm0.037$ dB. At seed 0, SSIM moves with it by only $0.001$ and LPIPS favors luminance-trained geometry on average; the gap does not depend on the appearance basis in the mean, and extending the solver budget changes the gap little on the two tested scenes.

The fixed-geometry recovery experiments are important because they change what a failed colorization result means. An independent per-Gaussian fit initially leaves a visible chromatic deficit, but most of that deficit disappears once residuals are attributed jointly across overlapping Gaussians. A weak inverse operator can therefore make a fixed spatial representation look less capable than it is. Conversely, recovery above the trained checkpoint does not imply hidden chroma in geometry; it reflects a quadratic post-hoc appearance objective, closer to the PSNR metric than the training loss, and access to the target training images, including views that training withheld.

The densification intervention shows that part of the geometry asymmetry is algorithmic. Chroma produces weaker values of the stock absolute densification statistic, so it receives fewer primitives under the default threshold. Correcting that gate at the first event recovers about $40\%$ of the loss. Densification driven by a fixed gradient threshold is known to be sensitive to gradient magnitude \citep{rotabulo2024revising,ye2024absgs,mallick2024taming}; our result shows that the channel content of the loss moves that operating point. Starting both arms from the same RGB-placed dense set with densification off recovers a further $0.6$--$1.1$ dB, a step that changes the initialization and the densifier together, and leaves $1.4$--$2.5$ dB with identical primitives. That set was placed under RGB supervision, so this remainder holds under a shared RGB-placed initialization; it is a property of the chromatic signal under the tested loss and optimizer rather than of the densifier. Separately, equalizing chromatic contrast in the loss recovers at most $9\%$ of the threshold-matched residual; the two controls were not combined. The channel hierarchy is therefore a property of image statistics interacting with the 3DGS objective, not of the densifier alone.

These findings suggest three possible changes to 3DGS, which we have not evaluated. Geometry-oriented operations such as densification could use luminance-weighted signals; chromatic appearance could be fitted or replaced after spatial support is established; and chromatic SH degree or precision could be allocated more aggressively than luminance. These observations are consistent with existing monochromatic reconstruction \citep{cheng2024colorizing,dhiman2025chromadistill,color3d2026}, fixed-geometry colorization \citep{lieber2026instant}, and YUV-domain compression methods \citep{gsico2026,higs2026,sony2026decorrelation}. Our experiments measure the conditions that make this division of labor possible and show that the asymmetry also appears during training. This motivates training procedures that treat luminance and chroma differently.

First, luminance is not universally sufficient: RGB-trained geometry keeps replicated advantages on \textsc{kitchen}, \textsc{dr.~johnson}, and \textsc{playroom}, and where a boundary is carried only by chroma, luminance-only training blurs its depth. Second, chroma is not geometrically irrelevant: conflicting chroma costs $0.15$--$0.67$ dB of geometry-supported reconstruction, replicated across seeds that each carry their own donor mapping. Third, chromatic appearance is not inherently view independent; on the chrome ball its relative view dependence matches that of luminance, even though its higher-order share of the recoverable span is smaller. Fourth, the RGB-versus-$Y$ comparison changes the color basis of the training loss together with the channel content; the basis control bounds this in the mean, but a loss-basis change cannot be removed without moving the densification operating point. Fifth, the causal channel study covers three scenes, with four seeds for the adversarial arm (each with its own donor mapping) and one seed for the chroma-only, threshold-matched, densification-free, contrast-stretched, and fork arms, and the opponent-basis sweep uses one seed per cell.
Sixth, the direct visible-geometry analysis is relative to retraining variation rather than ground-truth surfaces. In the object region of an analytic room with exact depth, luminance-only depth error matches RGB within $2\%$ on diffuse and colored-specular materials, but is much larger where luminance is flat. The prespecified criterion uses the whole image, where grazing-angle walls dominate and RGB itself fails the criterion. The object-region results are therefore descriptive. Testing a criterion fixed in advance for that region, with fresh seeds, is left to future work. Seventh, all conclusions remain specific to the tested 3DGS regimes; sparse-view settings, NeRFs, and other primitive families are natural next tests. 

Eighth, capture and image processing can alter chroma before training. Demosaicing and in-camera processing can smooth chromatic detail. The inspected benchmark JPEGs use 4:2:0 chroma subsampling, which stores chroma at half the luminance resolution in each dimension. Their near-lossless quantization tables do not establish that subsampling was the only loss of chromatic detail, or reveal earlier processing and compression. Training loads the Mip-NeRF 360 images downscaled by $1.95$--$3.24\times$ (Sec.~\ref{sec:experimental-framework}), which brings the stored chroma grid close to the training grid but cannot restore detail already lost, whereas the Tanks and Temples and Deep Blending scenes train at native resolution, with chroma at half the training resolution; the three scenes of the channel study are Mip-NeRF 360 scenes. The chroma-only deficit is therefore a property of captured chroma under this pipeline; a controlled capture or render without chroma subsampling could reduce it, and we have not tested that. Saving the existing JPEGs in a lossless format would not restore the lost detail. 

Ninth, every model starts from SfM points recovered from grayscale features, and opacity is frozen inside $G$. The conclusions concern the reconstruction a representation supports after standard optimization and densification from that initialization, and opacity can retain photometric structure that the appearance solve does not reset; re-solving opacity would require a nonlinear solve, which we do not attempt.

\section{Conclusion}
\label{sec:conclusion}

Standard RGB 3DGS training entangles the signals that build spatial support with the coefficients that color it. Separating those roles reveals a consistent but non-absolute asymmetry. Within standard SfM-initialized 3DGS, luminance provides the dominant signal for forming geometry that enables reconstruction and carries most of the benefit of higher-order directional appearance. Chroma contributes a weaker geometry signal, can measurably perturb optimization, and sometimes matters early, but much of its appearance can be re-estimated later on fixed support with low angular capacity. This division of labor explains why delayed chroma can often preserve final quality and suggests that future 3DGS systems need not allocate identical training and representation capacity to luminance and chroma.

\clearpage
\appendix
\section[Supplementary Material]{Supplementary Material}
\label{sec:supplementary}

\subsection{Appearance-Solver Validation}
The coupled solver uses visibility-weighted per-Gaussian normal equations followed by repeated render--residual--projection refinement. The ridge profile multiplies SH degrees $0,1,2,3$ by $1,10,100,1000$, respectively. A pre-primary sweep on \textsc{garden} and \textsc{kitchen} selected base ridge $10^{-6}$; recovery decreased with stronger regularization. That sweep, and the choice of iteration budgets from the quality curves, scored candidates on the benchmark test views of those two scenes; the settings were then fixed for the main geometry comparisons; the operator ladder and the capacity sweep use the separate settings reported below. The appearance fit itself never uses test images. The selection targeted chroma recovery on RGB-trained geometry and was never applied to luminance-trained geometry, so it can make the RGB arm's absolute recovery values optimistic; in expectation this works against luminance-trained geometry in the between-arm comparison, but the realized effect of the selection on that comparison was not measured. Both arms use identical settings. Production full-appearance solves use tolerance $3\times10^{-4}$ and a 60-iteration cap. All 22 base two-arm solves reach the cap symmetrically. Running the canonical solve to 200 iterations with every stopping test disabled, on \textsc{kitchen} and \textsc{counter}, moves the gap from $-0.4519$ to $-0.4657$ dB and from $+0.0352$ to $+0.0251$ dB; each arm gains $0.05$--$0.07$ dB over the extra 140 iterations, and they gain together, so the drift is $11$--$15\times$ smaller than the decision band. Iteration 60 reproduces the published rows to $4\times10^{-6}$ dB. An earlier bound of $0.0035$ dB, extrapolated from training residuals, understated this drift by about $4\times$; we quote the measurement. The operator-ladder figure used the earlier validation defaults for its coupled solves (ridge $10^{-5}$, tolerance $2\times10^{-3}$, 30-iteration cap); changing the ridge to $10^{-6}$ changes recovery by $0.0004$ on \textsc{garden} and $0.005$ on \textsc{kitchen}, so it does not alter the interpretation; these settings are why the ladder's coupled-DC value ($0.991$) and the capacity sweep's degree-0 chroma value ($0.994$) differ.

The working-space transform is not exactly render-invariant because the reference rasterizer clamps evaluated colors before compositing. Across validation scenes, the RGB-render/transform path and direct-YCbCr-render path differ by less than $0.11$ dB. Every fixed-geometry comparison uses the same direct working-space path for both arms. This standardizes evaluation, but geometry-dependent clamping effects have not been ruled out.

\subsection{Geometry Replication, Basis Control, and Direct Diagnostics}
The main comparison was first run at seed 0, which gave an eleven-scene mean of $-0.098\pm0.042$ dB with nine scenes inside the $\pm0.15$ dB band. The two seed-0 scenes outside the band, the three scenes within $0.03$ dB of its edge, and \textsc{counter} as an in-band control were then replicated with four independent training realizations per condition, and the remaining scenes were brought to four realizations afterward. The intermediate mixed-replication value, with four runs on six scenes, three on \textsc{garden} and \textsc{truck}, and one on \textsc{bicycle}, \textsc{room}, and \textsc{stump}, was $-0.094\pm0.033$ dB; completing the replication moved it by $0.009$ dB. The replicated condition-mean differences are $-0.284\pm0.103$ dB on \textsc{kitchen}, $-0.215\pm0.041$ dB on \textsc{dr.~johnson}, $-0.226\pm0.054$ dB on \textsc{playroom}, $-0.110\pm0.035$ dB on \textsc{bonsai}, $-0.102\pm0.051$ dB on \textsc{train}, and $+0.013\pm0.014$ dB on \textsc{counter}, and $+0.023\pm0.008$ dB on \textsc{bicycle}, $-0.060\pm0.009$ dB on \textsc{garden}, $-0.003\pm0.059$ dB on \textsc{room}, $+0.119\pm0.042$ dB on \textsc{stump}, and $-0.085\pm0.027$ dB on \textsc{truck} (Table~\ref{tab:four-run-per-scene}). \textsc{playroom} moves from $-0.115$ dB at one seed to outside the band; \textsc{bonsai} and \textsc{train} stay inside. The eleven-scene mean with four runs on every scene is $-0.085$ dB (SE $0.037$; 95\% CI $[-0.166,-0.003]$). Equivalence at $\pm0.15$ dB is not established post hoc, because the 90\% interval misses the band by $0.001$ dB, and the band was not widened. Runs with the same seed label share the camera order, so a paired analysis is also available; GPU rasterization is nondeterministic, so the pairing is partial. Treating the four same-seed-label differences per scene as paired gives per-scene standard errors of $0.006$--$0.117$ dB (median $0.029$) against $0.008$--$0.103$ dB for the unpaired calculation in Table~\ref{tab:four-run-per-scene}, and leaves the eleven-scene mean, its between-scene interval, and the three departures unchanged.

\begin{table}[t]
\centering
\small
\color{black}
\caption{\textbf{Four-run main comparison per scene.} Condition means of held-out PSNR after $A^*$ over four independent training runs per condition. $\Delta_{\mathrm{geom}}$ is $Y$ minus RGB; its standard error is $\sqrt{s^2_{\mathrm{RGB}}/4+s^2_{Y}/4}$ from the per-run values. A scene is a departure when $|\Delta_{\mathrm{geom}}|>0.15$ dB; all three departures also exceed twice their standard error. The standard error in the mean row is the between-scene standard deviation divided by $\sqrt{11}$.}
\label{tab:four-run-per-scene}
\begin{tabular}{@{}lrrrrl@{}}
\toprule
Scene & $A^*(G_{\mathrm{RGB}})$ & $A^*(G_Y)$ & $\Delta_{\mathrm{geom}}$ & SE & Status\\
\midrule
bicycle & 25.409 & 25.432 & $+0.023$ & 0.008 & in band\\
bonsai & 33.133 & 33.023 & $-0.110$ & 0.035 & in band\\
counter & 29.621 & 29.634 & $+0.013$ & 0.014 & in band\\
garden & 27.660 & 27.600 & $-0.060$ & 0.009 & in band\\
kitchen & 32.056 & 31.772 & $-0.284$ & 0.103 & departure\\
room & 32.185 & 32.182 & $-0.003$ & 0.059 & in band\\
stump & 26.515 & 26.635 & $+0.119$ & 0.042 & in band\\
train & 22.395 & 22.293 & $-0.102$ & 0.051 & in band\\
truck & 25.860 & 25.775 & $-0.085$ & 0.027 & in band\\
playroom & 29.832 & 29.606 & $-0.226$ & 0.054 & departure\\
drjohnson & 29.359 & 29.144 & $-0.215$ & 0.041 & departure\\
\midrule
Mean & 28.548 & 28.463 & $-0.085$ & 0.037 & 8 of 11 in band\\
\bottomrule
\end{tabular}
\end{table}

The basis control trains the YCbCr SH parameterization under the stock RGB loss. Its Gaussian counts are $0.967$--$1.020$ of the RGB count on all eleven scenes, and its matched-appearance difference from RGB training is $-0.018$ dB (SE $0.041$; 95\% CI $[-0.109,+0.073]$), with eight of eleven scenes inside the band at a single seed (\textsc{kitchen} moves in; \textsc{train} and \textsc{playroom} move out, \textsc{playroom} by $0.007$ dB). A naive full-YCbCr loss arm, which averages the stock loss over the three transformed channels, densified to $14\%$--$40\%$ of the RGB count on all eleven scenes and was retained only as a diagnostic; on \textsc{counter} its reconstruction ($29.121$ dB) matches the reduced-luminance control ($29.037$ dB) within $0.15$ dB, consistent with a deficit caused partly by fewer Gaussians. The mechanism is that averaging the three transformed channels changes both the scale and the structure of the image-space gradients, so the fixed densification threshold no longer represents the same operating point.

Four scenes trigger the primitive-count control at seed 0. The $Y$ surplus relative to the RGB count is $6.1\%$ on \textsc{counter}, $6.6\%$ on \textsc{garden}, $9.7\%$ on \textsc{stump}, and $6.3\%$ on \textsc{playroom}. After removing the lowest-opacity $Y$ Gaussians to match the RGB count and re-solving $A^*$, held-out PSNR changes by at most $0.005$ dB and the RGB-vs-$Y$ gap by at most $0.006$ dB.

Rendered expected depth uses front-to-back weights $w_{i,p}=\alpha_{i,p}T_{i,p}$ and is compared through median absolute log-depth difference on common visible support. On replicated in-band scenes \textsc{counter} and \textsc{truck}, cross-condition differences are on the same scale as within-RGB and within-$Y$ seed variation. \textsc{kitchen} has a modest but elevated shift relative to its own seed null, while \textsc{dr.~johnson} differs more clearly in visible support. Across the eleven seed-0 scenes, expected-depth discrepancy does not rank with the common-appearance reconstruction deficit ($\rho=0.064$). These diagnostics therefore bound the claim: $A^*$ measures reconstruction support, not physical-surface identity.

\subsection{Protocol Amendments}
Three departures from the prespecified protocol were made before the results were finalized, and we record them here. First, the original rule excluded solves that reached the iteration cap without meeting the stopping tolerance. Every solve of the main RGB-versus-$Y$ comparison reached the cap, in both arms alike, so the rule was amended to retain capped solves and the 200-iteration convergence check of Sec.~\ref{sec:supplementary} was added in its place. Second, the Gaussian-count control was planned as retraining at a matched count. It was replaced by the opacity-pruning diagnostic reported above, which removes the lowest-opacity Gaussians from the larger model and re-solves $A^*$; this is cheaper and holds the trained geometry fixed, but it does not test what training would have produced at the matched count. Third, the adversarial-chroma arm was first run with donor views drawn from all views of each scene, including benchmark test views and validation views, and with one mapping shared by all seeds. It was rerun with donors restricted to the optimization views and a fresh mapping per seed, and only the rerun is reported. The per-scene deficits moved by at most $0.10$ dB between the two runs and every gate outcome was unchanged.

\subsection[SSIM and LPIPS Companion Tables]{SSIM and LPIPS Companion Tables}
{\color{black}
Tables~\ref{tab:solver-geometry-ssim-lpips} and~\ref{tab:capacity-ssim-lpips} give SSIM and LPIPS for the main geometry comparison, per scene, and for the SH-capacity sweep. Both come from the same solves and evaluation path as the PSNR values in the main text.\par}

\begin{table*}[t]
\centering
\small
\color{black}
\caption{\textbf{SSIM and LPIPS per scene for the main comparison.} Seed 0, with the same rows and evaluation path as Table~\ref{tab:solver-geometry}; $Q_0$ is evaluated through the common YCbCr render-to-RGB path. SSIM is higher-better and LPIPS lower-better, so a positive LPIPS $Q_2-Q_1$ favors RGB geometry.}
\label{tab:solver-geometry-ssim-lpips}
\begin{tabular}{@{}lrrrrrrrr@{}}
\toprule
& \multicolumn{4}{c}{SSIM $\uparrow$} & \multicolumn{4}{c}{LPIPS $\downarrow$} \\
\cmidrule(lr){2-5}\cmidrule(lr){6-9}
Scene & $Q_0$ & $Q_1$ & $Q_2$ & $Q_2-Q_1$ & $Q_0$ & $Q_1$ & $Q_2$ & $Q_2-Q_1$ \\
\midrule
bicycle & 0.7427 & 0.7429 & 0.7449 & $+0.0020$ & 0.2094 & 0.2113 & 0.2079 & $-0.0034$ \\
bonsai & 0.9427 & 0.9459 & 0.9447 & $-0.0011$ & 0.1262 & 0.1225 & 0.1221 & $-0.0005$ \\
counter & 0.9068 & 0.9130 & 0.9129 & $-0.0000$ & 0.1516 & 0.1463 & 0.1444 & $-0.0020$ \\
garden & 0.8554 & 0.8546 & 0.8536 & $-0.0009$ & 0.0903 & 0.0946 & 0.0932 & $-0.0013$ \\
kitchen & 0.9300 & 0.9329 & 0.9310 & $-0.0020$ & 0.0885 & 0.0893 & 0.0893 & $-0.0001$ \\
room & 0.9214 & 0.9272 & 0.9274 & $+0.0002$ & 0.1607 & 0.1593 & 0.1585 & $-0.0008$ \\
stump & 0.7567 & 0.7619 & 0.7643 & $+0.0024$ & 0.1922 & 0.1899 & 0.1809 & $-0.0090$ \\
train & 0.8186 & 0.8173 & 0.8140 & $-0.0033$ & 0.1586 & 0.1587 & 0.1602 & $+0.0015$ \\
truck & 0.8829 & 0.8824 & 0.8810 & $-0.0014$ & 0.0987 & 0.0993 & 0.0985 & $-0.0008$ \\
playroom & 0.9045 & 0.9070 & 0.9016 & $-0.0054$ & 0.1624 & 0.1663 & 0.1684 & $+0.0021$ \\
drjohnson & 0.9020 & 0.9035 & 0.9009 & $-0.0026$ & 0.1774 & 0.1787 & 0.1803 & $+0.0016$ \\
\midrule
Mean & 0.8694 & 0.8717 & 0.8706 & $-0.0011$ & 0.1469 & 0.1469 & 0.1458 & $-0.0011$ \\
\bottomrule
\end{tabular}
\end{table*}

\begin{table}[t]
\centering
\small
\color{black}
\caption{\textbf{Channel capacity in PSNR, SSIM, and LPIPS.} Eleven-scene means of the full-appearance solve restricted to SH degree $\le$ cap for one channel group; the other channel group keeps its trained appearance, and this sweep uses tolerance $5\times10^{-4}$ and a 70-iteration cap, so its degree-3 rows are not the $A^*$ rows of Table~\ref{tab:solver-geometry}.}
\label{tab:capacity-ssim-lpips}
\begin{tabular}{@{}llrrr@{}}
\toprule
Group & Cap & PSNR $\uparrow$ & SSIM $\uparrow$ & LPIPS $\downarrow$ \\
\midrule
Y & 0 & 26.925 & 0.8392 & 0.1683 \\
 & 1 & 28.059 & 0.8615 & 0.1541 \\
 & 2 & 28.291 & 0.8661 & 0.1518 \\
 & 3 & 28.366 & 0.8673 & 0.1514 \\
\midrule
CbCr & 0 & 28.049 & 0.8679 & 0.1484 \\
 & 1 & 28.206 & 0.8711 & 0.1445 \\
 & 2 & 28.229 & 0.8717 & 0.1441 \\
 & 3 & 28.236 & 0.8718 & 0.1440 \\
\bottomrule
\end{tabular}
\end{table}

\subsection[Per-Scene Solver Comparison and MSE-Domain Recovery]{Per-Scene Solver Comparison and MSE-Domain Recovery}
Table~\ref{tab:per-scene-q012} gives $Q_0$, $Q_1$, and $Q_2$ per scene at seed 0, with the pooled held-out MSE of the trained reference; pooled MSE was not recorded for the seed-0 $A^*$ solves. Table~\ref{tab:recovery-ladder} gives the per-scene PSNR-domain recovery of the four operator-ladder solves with the denominators of both ratios. The ladder solves predate per-view logging, so their MSE-domain ratios are not available; their trained-reference and chroma-removed states are identical to those of the capacity sweep (to $10^{-6}$ dB on every scene), so the MSE denominators are exact. Table~\ref{tab:recovery-mse} gives both ratios for the capacity sweep. In the MSE domain the ordering of the two channel groups is unchanged at every cap, and mean recovery is non-decreasing in the cap for both groups.

\begin{table}[t]
\centering
\small
\color{black}
\caption{\textbf{Per-scene $Q_0$, $Q_1$, and $Q_2$ at seed 0.} PSNR in dB on the benchmark test views. $Q_0$ is the trained RGB appearance evaluated through the common YCbCr render-to-RGB path; $Q_1$ and $Q_2$ are $A^*$ on $G_{\mathrm{RGB}}$ and $G_Y$. The last column is the pooled held-out MSE of the trained reference; pooled MSE was not recorded for the seed-0 $A^*$ solves.}
\label{tab:per-scene-q012}
\begin{tabular}{@{}lrrrrrr@{}}
\toprule
Scene & $Q_0$ & $Q_1$ & $Q_2$ & $Q_1-Q_0$ & $Q_2-Q_1$ & MSE($Q_0$) $\times10^{-3}$\\
\midrule
bicycle & 25.070 & 25.411 & 25.448 & $+0.341$ & $+0.037$ & 3.836\\
bonsai & 32.376 & 33.192 & 33.070 & $+0.817$ & $-0.123$ & 0.7725\\
counter & 28.752 & 29.628 & 29.663 & $+0.876$ & $+0.035$ & 1.486\\
garden & 27.261 & 27.649 & 27.609 & $+0.387$ & $-0.039$ & 2.254\\
kitchen & 31.542 & 32.166 & 31.714 & $+0.624$ & $-0.452$ & 0.7777\\
room & 31.673 & 32.156 & 32.089 & $+0.482$ & $-0.067$ & 0.8860\\
stump & 26.389 & 26.574 & 26.614 & $+0.185$ & $+0.040$ & 2.605\\
train & 21.834 & 22.346 & 22.225 & $+0.513$ & $-0.121$ & 7.763\\
truck & 25.348 & 25.864 & 25.771 & $+0.516$ & $-0.093$ & 3.020\\
playroom & 29.851 & 29.810 & 29.695 & $-0.041$ & $-0.115$ & 1.315\\
drjohnson & 29.136 & 29.311 & 29.130 & $+0.175$ & $-0.181$ & 1.733\\
\midrule
Mean & 28.112 & 28.555 & 28.457 & $+0.443$ & $-0.098$ & 2.404\\
\bottomrule
\end{tabular}
\end{table}

\begin{table}[t]
\centering
\footnotesize
\setlength{\tabcolsep}{3.5pt}
\color{black}
\caption{\textbf{Operator ladder per scene, with the denominators of both recovery ratios.} $R_{CbCr}$ is the PSNR-domain ratio of Eq.~\eqref{eq:recovery-compact} for the four solves of Fig.~\ref{fig:geometry-and-recovery}B. The PSNR denominator is $Q(A_{\mathrm{ref}})-Q(A^{\varnothing}_{CbCr})$. The MSE columns give the pooled held-out MSE of the trained reference and of the chroma-removed state, and their difference, which is the denominator of $R^{\mathrm{MSE}}$.}
\label{tab:recovery-ladder}
\begin{tabular}{@{}lrrrrrrrr@{}}
\toprule
& \multicolumn{4}{c}{$R_{CbCr}$ (PSNR domain)} & PSNR den. & \multicolumn{3}{c}{MSE ($\times10^{-3}$)}\\
\cmidrule(lr){2-5}\cmidrule(lr){7-9}
Scene & tint & indep.\ DC & coupled DC & coupled d3 & (dB) & ref. & removed & den.\\
\midrule
bicycle & 0.427 & 0.956 & 0.990 & 1.009 & 3.096 & 3.836 & 6.741 & 2.906\\
bonsai & 0.007 & 0.858 & 0.971 & 1.022 & 7.884 & 0.7725 & 3.703 & 2.931\\
counter & 0.231 & 0.942 & 1.007 & 1.036 & 9.860 & 1.486 & 13.28 & 11.79\\
garden & 0.458 & 0.920 & 0.986 & 1.007 & 8.051 & 2.254 & 12.14 & 9.885\\
kitchen & 0.137 & 0.814 & 0.946 & 1.010 & 12.993 & 0.7777 & 14.13 & 13.36\\
room & 0.272 & 0.912 & 0.994 & 1.009 & 11.745 & 0.8860 & 10.78 & 9.893\\
stump & 0.695 & 0.944 & 0.996 & 1.011 & 5.228 & 2.605 & 7.781 & 5.177\\
train & 0.042 & 0.856 & 1.001 & 1.042 & 3.220 & 7.763 & 14.94 & 7.173\\
truck & 0.218 & 0.949 & 1.004 & 1.019 & 5.725 & 3.020 & 11.39 & 8.374\\
playroom & 0.495 & 0.965 & 1.003 & 1.008 & 16.473 & 1.315 & 48.06 & 46.75\\
drjohnson & 0.322 & 0.940 & 1.001 & 1.010 & 5.607 & 1.733 & 4.751 & 3.018\\
\midrule
Mean & 0.300 & 0.914 & 0.991 & 1.017 & 8.171 & 2.404 & 13.43 & 11.02\\
\bottomrule
\end{tabular}
\end{table}

\begin{table}[t]
\centering
\footnotesize
\setlength{\tabcolsep}{3.5pt}
\color{black}
\caption{\textbf{Capacity sweep in both domains.} Recovery when one channel group is re-solved through SH degree $\le$ cap while the other keeps its trained appearance. $R$ is the PSNR-domain ratio of Eq.~\eqref{eq:recovery-compact} and $R^{\mathrm{MSE}}$ its MSE-domain counterpart (Sec.~\ref{sec:experimental-framework}), each with its denominator. Solver settings: ridge $10^{-6}$, tolerance $5\times10^{-4}$, 70-iteration cap.}
\label{tab:recovery-mse}
\begin{tabular}{@{}lrrrrrrrrrr@{}}
\toprule
& \multicolumn{4}{c}{$R$ at cap} & \multicolumn{4}{c}{$R^{\mathrm{MSE}}$ at cap} & PSNR den. & MSE den.\\
\cmidrule(lr){2-5}\cmidrule(lr){6-9}
Scene & 0 & 1 & 2 & 3 & 0 & 1 & 2 & 3 & (dB) & ($\times10^{-3}$)\\
\midrule
\multicolumn{11}{@{}l}{\emph{Luminance ($Y$)}}\\
bicycle & 0.928 & 1.009 & 1.024 & 1.027 & 0.987 & 1.003 & 1.005 & 1.006 & 11.635 & 42.71\\
bonsai & 0.872 & 0.985 & 1.014 & 1.024 & 0.994 & 1.000 & 1.001 & 1.002 & 21.478 & 82.30\\
counter & 0.882 & 0.988 & 1.015 & 1.023 & 0.989 & 0.999 & 1.001 & 1.002 & 17.554 & 74.97\\
garden & 0.900 & 0.998 & 1.018 & 1.024 & 0.983 & 1.000 & 1.003 & 1.004 & 12.432 & 31.23\\
kitchen & 0.875 & 0.991 & 1.016 & 1.022 & 0.986 & 0.999 & 1.001 & 1.002 & 16.100 & 28.43\\
room & 0.956 & 1.006 & 1.012 & 1.014 & 0.998 & 1.000 & 1.000 & 1.000 & 20.364 & 76.79\\
stump & 0.946 & 0.995 & 1.004 & 1.007 & 0.993 & 0.999 & 1.000 & 1.000 & 13.845 & 54.04\\
train & 0.931 & 1.004 & 1.018 & 1.025 & 0.980 & 0.998 & 1.004 & 1.006 & 10.396 & 66.94\\
truck & 0.938 & 1.008 & 1.023 & 1.027 & 0.989 & 1.002 & 1.004 & 1.005 & 12.481 & 49.86\\
playroom & 0.959 & 0.982 & 0.985 & 0.987 & 0.996 & 0.997 & 0.996 & 0.997 & 15.955 & 43.42\\
drjohnson & 0.976 & 1.007 & 1.008 & 1.009 & 0.999 & 1.002 & 1.002 & 1.002 & 15.754 & 49.52\\
Mean & 0.924 & 0.998 & 1.012 & 1.017 & 0.990 & 1.000 & 1.002 & 1.002 & 15.272 & 54.56\\
\addlinespace
\multicolumn{11}{@{}l}{\emph{Chroma ($CbCr$)}}\\
bicycle & 0.993 & 1.007 & 1.007 & 1.008 & 0.995 & 1.004 & 1.004 & 1.004 & 3.096 & 2.906\\
bonsai & 0.978 & 1.018 & 1.027 & 1.030 & 0.997 & 1.010 & 1.014 & 1.015 & 7.884 & 2.931\\
counter & 1.011 & 1.034 & 1.038 & 1.039 & 1.007 & 1.012 & 1.013 & 1.013 & 9.860 & 11.79\\
garden & 0.989 & 1.006 & 1.008 & 1.008 & 0.997 & 1.001 & 1.002 & 1.002 & 8.051 & 9.885\\
kitchen & 0.954 & 1.003 & 1.014 & 1.017 & 0.992 & 1.001 & 1.002 & 1.003 & 12.993 & 13.36\\
room & 0.997 & 1.008 & 1.010 & 1.011 & 0.999 & 1.001 & 1.001 & 1.001 & 11.745 & 9.893\\
stump & 0.998 & 1.005 & 1.008 & 1.008 & 0.999 & 1.000 & 1.002 & 1.003 & 5.228 & 5.177\\
train & 1.006 & 1.040 & 1.040 & 1.039 & 1.003 & 1.024 & 1.021 & 1.020 & 3.220 & 7.173\\
truck & 1.007 & 1.019 & 1.021 & 1.021 & 1.003 & 1.008 & 1.009 & 1.009 & 5.725 & 8.374\\
playroom & 1.003 & 1.005 & 1.002 & 1.001 & 1.000 & 1.000 & 1.000 & 1.000 & 16.473 & 46.75\\
drjohnson & 1.002 & 1.000 & 1.000 & 1.000 & 1.004 & 1.000 & 1.000 & 1.001 & 5.607 & 3.018\\
Mean & 0.994 & 1.013 & 1.016 & 1.017 & 1.000 & 1.006 & 1.006 & 1.007 & 8.171 & 11.02\\
\bottomrule
\end{tabular}
\end{table}

\subsection{Recovery Robustness, Specularity, and Generality}
\label{sec:robustness-generality}
Coupled degree-3 chroma recovery remains near the trained reference across post-release checkpoints and independent official Inria checkpoints. At 30k, delayed-chroma checkpoints average approximately $1.002$ recovery; official reference checkpoints average approximately $1.012$. Extending \textsc{kitchen} from 30k to 60k changes recovery from $1.005$ to $1.002$. These tests show that near-complete recoverability is not specific to one checkpoint time or our training pipeline.

The SH-capacity result is also tested on shiny scenes. Higher-order luminance remains more valuable in every tested case, while chromatic higher-order cost becomes measurable when reflected chromaticity changes with viewpoint. On the chrome ball, removing higher-order chroma costs about $0.61$ dB and removing higher-order luminance about $5.7$ dB. In span-normalized terms, degrees 1--3 account for a mean $0.116$ of luminance's recoverable span over nine specular scenes against $0.034$ for chroma over eight (\textsc{gardenspheres} chroma excluded because its recoverable span is below $1$ dB), a gap of $+0.082$ against a frozen threshold of $0.0355$; on the matched eight scenes the luminance share is $0.134$ and the gap $+0.100$; the gap is $+0.115$ without the two scenes that disagree with the collapse test and $+0.056$ without the chrome ball. A fixed-surface-point analysis of ground-truth views shows that the across-view spread of $Cb$ and $Cr$ is roughly $57\%$ and $62\%$ of their spatial spread within a view, against $47\%$ for $Y$, confirming that the small absolute chromatic cost does not imply absence of directional chromatic structure.

Delayed-chroma training transfers qualitatively to FastGS. Alternative affine opponent bases (BT.601, BT.709, YCoCg, and YIQ) also preserve the same broad release-and-recovery pattern: final parity and release-jump gates pass in 40 of 44 scene/basis cells, with both passing in 36 of 44. These controls argue against an effect tied only to one 3DGS implementation or one luminance matrix. Nonlinear perceptual spaces were tested only post hoc and are not treated as equivalent training bases.

\subsection{Delayed-Chroma Boundary Controls}
\begin{figure}[h!]
    \centering

    \begin{minipage}[t]{0.49\textwidth}
        \centering
        \includegraphics[width=\linewidth]
            {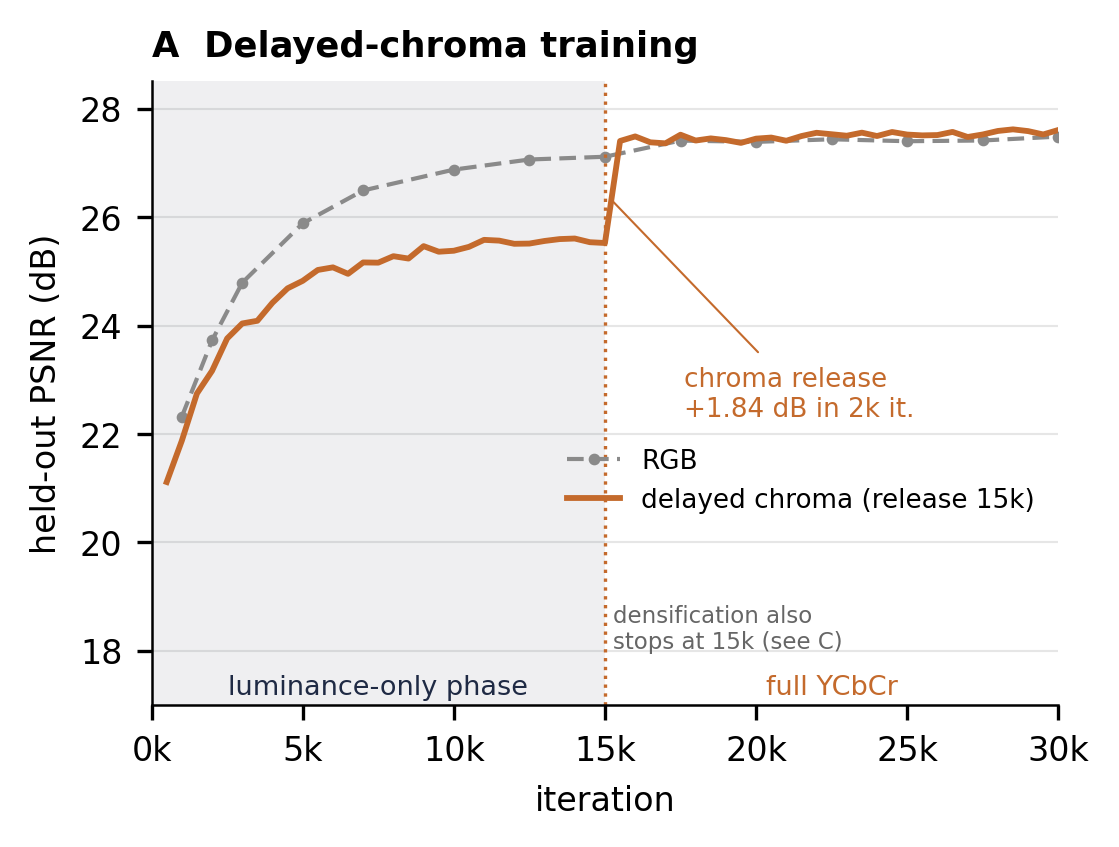}
    \end{minipage}
    \hfill
    \begin{minipage}[t]{0.49\textwidth}
        \centering
        \includegraphics[width=\linewidth]
            {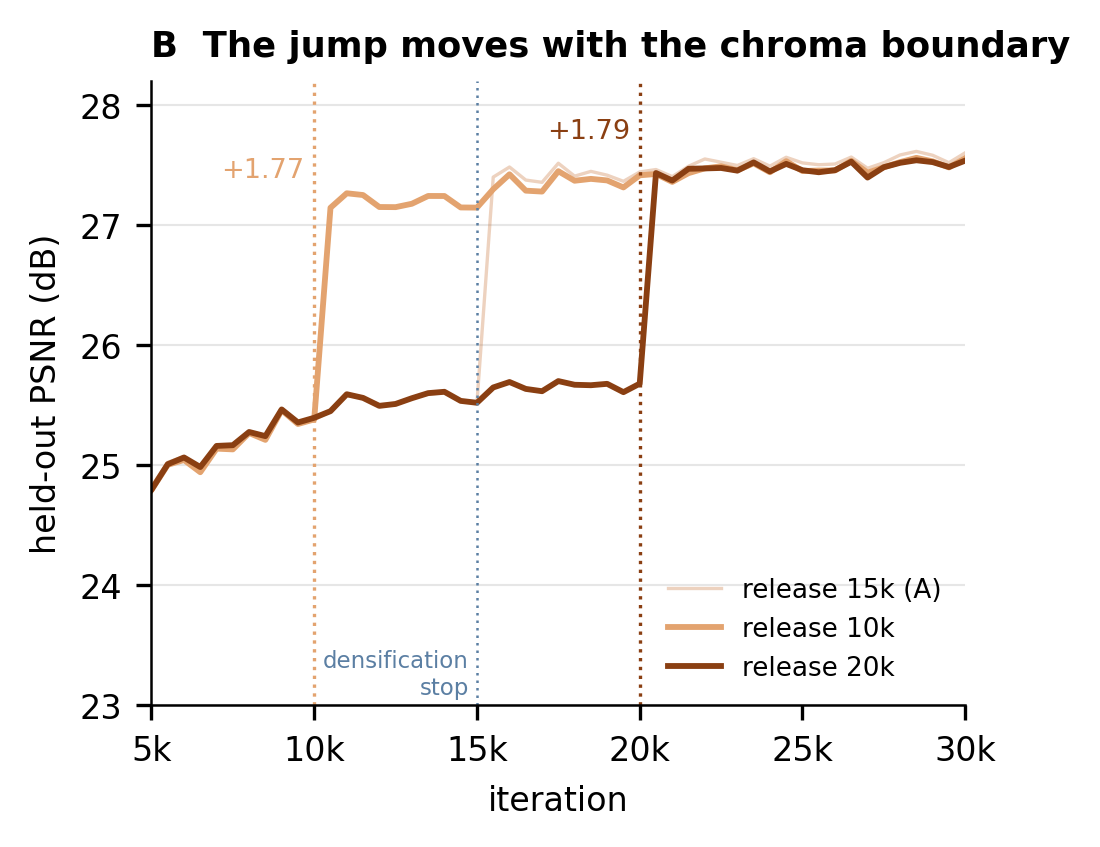}
    \end{minipage}

    \vspace{0.8em}

    \begin{minipage}[t]{0.49\textwidth}
        \centering
        \includegraphics[width=\linewidth]
            {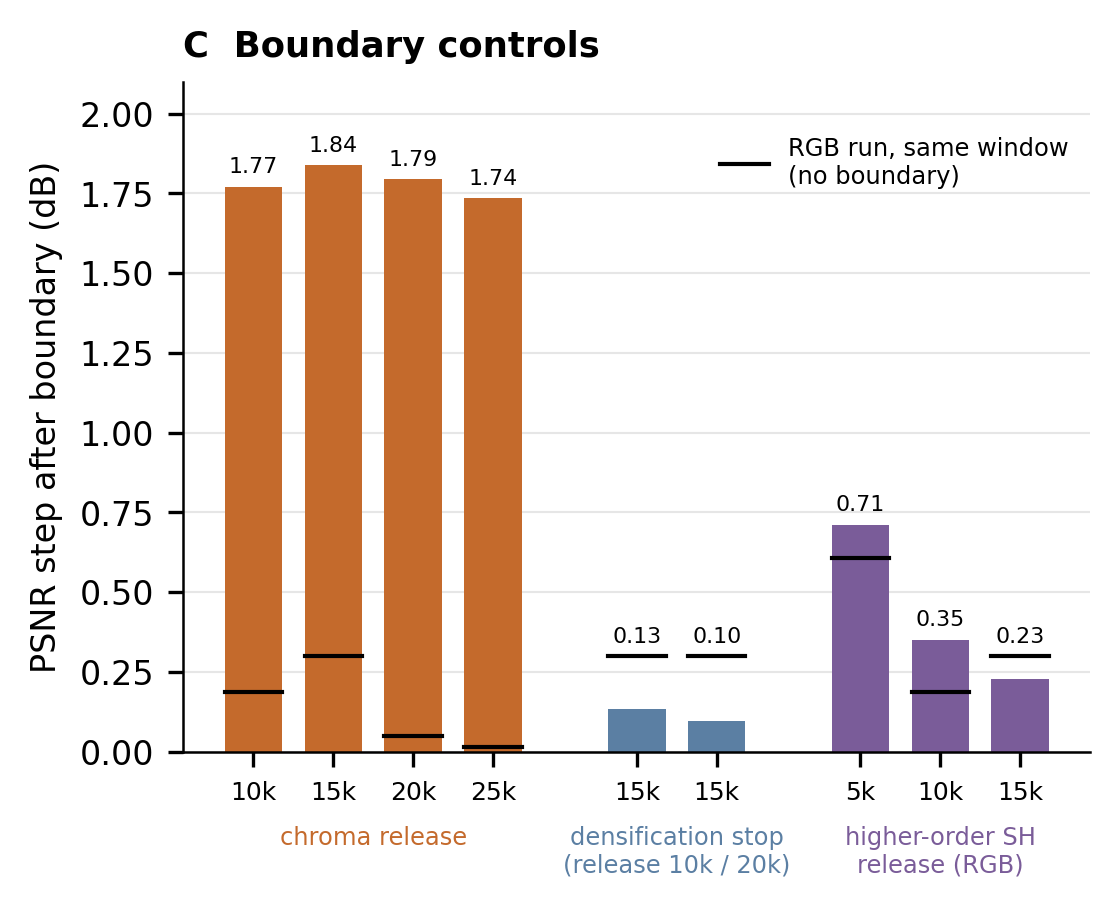}
    \end{minipage}
    \hfill
    \begin{minipage}[t]{0.49\textwidth}
        \centering
        \includegraphics[width=\linewidth]
            {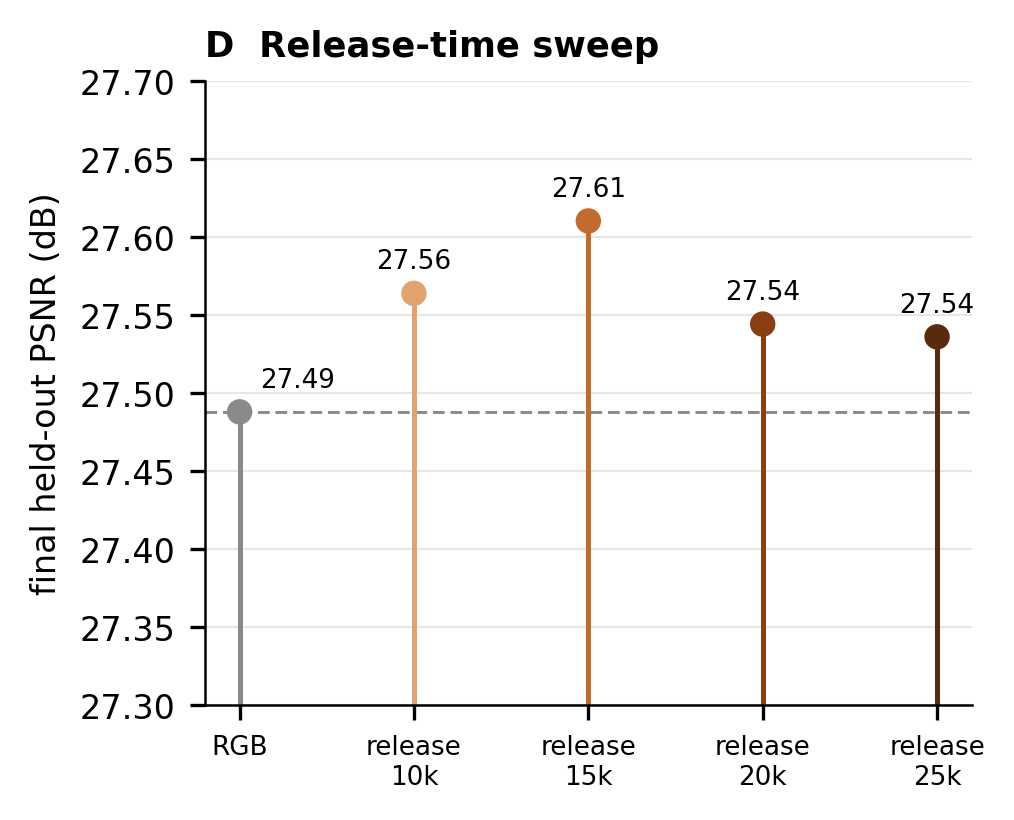}
    \end{minipage}

    \caption{
    \textbf{Delayed-chroma training dynamics and boundary controls in stock
    3DGS.}
    \textbf{(A)} Training begins with luminance-only supervision and switches
    to full YCbCr supervision at 15k iterations. Chroma release produces a
    \(1.84~\mathrm{dB}\) increase in held-out PSNR over the
    following 2k iterations, after which the delayed schedule converges near
    the RGB reference.
    \textbf{(B)} Moving chroma release moves the reconstruction jump with it:
    releases at 10k, 15k, and 20k produce approximately
    \(1.77\), \(1.84\), and \(1.79~\mathrm{dB}\) steps, respectively,
    despite the fixed densification boundary at 15k.
    \textbf{(C)} Boundary controls separate chroma release from coincident
    optimization events. Chroma release produces
    \(1.74\)--\(1.84~\mathrm{dB}\) changes across release times, whereas
    stopping densification produces only \(0.10\)--\(0.13~\mathrm{dB}\)
    changes. Each bar spans 2{,}000 iterations from its boundary, except the
    25k release, which spans 2{,}500. Releasing higher-order SH without changing
    chromatic supervision produces smaller and non-systematic effects. Black
    horizontal marks show the change in an RGB run from the same boundary;
    the RGB run was logged at 2{,}500-iteration spacing beyond 5k, so its
    marks span 2{,}500 iterations where the bars span 2{,}000.
    \textbf{(D)} The four delayed-release schedules finish within
    \(0.074~\mathrm{dB}\) of one another. Including the standard RGB
    reference, the full five-condition spread is \(0.122~\mathrm{dB}\).
    All panels use the stock 3DGS implementation; the independent FastGS
    transfer is reported in Sec.~\ref{sec:robustness-generality}.
    }
    \label{fig:training-dynamics}
\end{figure}
The stock-3DGS release sweep produces approximately $1.74$--$1.84$ dB steps when chroma is introduced between 10k and 25k iterations. Windows of 2{,}000 iterations around the end of densification show changes of only $0.10$--$0.13$ dB, and releasing higher-order SH without changing supervised channels produces smaller, non-systematic effects. On the \textsc{garden} release sweep, RGB training ends at $27.488$ dB and releases at 10k, 15k, 20k, and 25k end at $27.564$, $27.610$, $27.544$, and $27.536$ dB, respectively. The delayed schedules span only $0.074$ dB. Two early-freeze controls freeze the representation at the 15k release and train only chromatic appearance to 30k: one trains only the chromatic DC term, the other all chromatic SH. Relative to the delayed schedule that keeps everything trainable, scored on the eight validation views, they reduce final quality by $0.809$ and $0.602$ dB on average over the eleven scenes; per scene the costs run from $+0.096$ and $+0.124$ dB on \textsc{stump}, the only positive case, to $-1.721$ and $-1.561$ dB on \textsc{room}. Geometry and luminance appearance therefore continue to improve after chroma release, and the release experiment does not imply that geometry is complete at the boundary. The shared-checkpoint fork continues one luminance-only checkpoint per scene from 15k to 30k twice, without and with chromatic supervision; both branches carry the same set of primitives at every evaluation. The chroma branch changes $A^*$-supported reconstruction by $+0.113$, $+0.060$, and $+0.118$ dB on \textsc{garden}, \textsc{counter}, and \textsc{kitchen}, and its three-scene mean is $0.001$ dB above the three-scene mean of the four-run RGB arm, a difference of means rather than a per-scene bound (per scene $+0.014$, $+0.077$, and $-0.087$ dB). The mean change of $+0.097$ dB matches the RGB advantage of the main comparison. All three values are positive, but each comes from a single checkpoint, so these estimates are suggestive.

\subsection{Causal-Intervention Details}
At the first densification event, the chroma threshold is chosen so the fraction of $CbCr$ Gaussians exceeding it matches the $Y$ fraction above the stock threshold on the identical SfM population. The threshold is frozen before observing final outcome. Over the full densification window, cumulative chroma clone/split proposals increase from only a few percent of the $Y$ totals under the stock threshold to approximately $61\%$--$81\%$ under threshold matching, with no compensating pruning increase. Despite this large correction, the final common-$A^*$ deficit remains $2.3$--$3.2$ dB. The densification-free comparison starts both arms from a stock RGB run's point set at 15k and trains 30k iterations with densification, pruning, and opacity resets disabled; both arms end with exactly the initial count on every scene ($4{,}218{,}069$, $1{,}050{,}890$, and $1{,}530{,}247$), and the chroma-only deficit is $1.519$, $1.363$, and $2.519$ dB (mean $1.800$ dB). The raw, threshold-matched, and densification-free deficits are therefore $4.035$, $2.330$, and $1.519$ dB on \textsc{garden}; $3.865$, $2.472$, and $1.363$ dB on \textsc{counter}; and $5.540$, $3.160$, and $2.519$ dB on \textsc{kitchen}.

The contrast-stretched arm scales $Cb$ and $Cr$ to the standard deviation of luminance inside the loss, with factors frozen from the ground-truth images ($4.09/6.26$ on \textsc{garden}, $3.54/3.42$ on \textsc{counter}, $2.46/3.00$ on \textsc{kitchen}), and re-matches the densification threshold on a stretched calibration run because the stretch also scales the densification statistic. The statistic was amplified $6.1\times$, $4.7\times$, and $3.3\times$, and first-event crossing matched luminance's rate to within $0.2\%$. The gains over the threshold-matched arm are $+0.206$, $+0.072$, and $+0.091$ dB, closing $9\%$, $3\%$, and $3\%$ of the residual.

The loss-scaled arms of Fig.~\ref{fig:chroma-geometry-controls}a use a per-scene factor $s$, the ratio of the luminance-only to the chroma-only densification statistic (the median, over densification events between iterations 500 and 15k, of the per-event median accumulated view-space positional-gradient norm), frozen before this arm's outcome was inspected, with pilot results from earlier arms already available: $1/s=1.496$, $0.411$, and $0.653$ on \textsc{garden}, \textsc{counter}, and \textsc{kitchen}. Chroma-only with the loss multiplied by $s$ reaches $23.111$, $26.781$, and $26.833$ dB after $A^*$, against $23.535$, $25.776$, and $26.294$ dB raw. Luminance-only with the loss multiplied by $1/s$ reaches $27.706$, $29.037$, and $31.400$ dB, against $27.570$, $29.641$, and $31.833$ dB unscaled, and on \textsc{counter} ends with $357{,}252$ Gaussians against $1{,}113{,}801$; the densification threshold is unchanged in both arms, so the scaling acts through the densification statistic. On \textsc{garden} the frozen factor fell on the wrong side of one, so both loss-scaled arms move away from the intended operating point and are not valid controls for that scene.

The adversarial-chroma condition replaces each optimization view's chroma with that of another optimization view under a derangement, so no view keeps its own chroma and no benchmark test view or validation view contributes any; each seed draws its own derangement from its training seed, and the mappings are stored with the experiment record. Luminance is preserved to floating-point precision before quantization; the recorded uint8 conversion changes it by at most $0.5/255$, with no clipping. The swap reduces the Pearson correlation between luminance- and chroma-gradient magnitudes from $0.50$--$0.55$ to $0.17$--$0.22$. Across four seeds per arm on every scene, the common-$A^*$ deficits relative to the luminance-only reference are $-0.145\pm0.020$, $-0.248\pm0.037$, and $-0.665\pm0.097$ dB on \textsc{garden}, \textsc{counter}, and \textsc{kitchen}, and relative to the RGB reference $-0.206\pm0.021$, $-0.233\pm0.035$, and $-0.933\pm0.047$ dB, the last two at matched Gaussian count (raw: $-0.235\pm0.035$ and $-0.949\pm0.046$ dB; on \textsc{garden} the count ratio of $0.976$ was within the 5\% gate, so no matching was applied). Every adversarial seed scores below every reference seed of either arm; the smallest margins are $0.094$ dB on \textsc{garden}, $0.151$ dB on \textsc{counter}, and $0.407$ dB on \textsc{kitchen}.
Pruning the RGB reference by lowest opacity to the adversarial count costs $0.002$ dB on \textsc{counter} (about $10\%$ pruned) and $0.016$ dB on \textsc{kitchen} (about $20\%$), so matched and raw comparisons agree within $0.016$ dB. Because each seed carries its own mapping, the adversarial arm's seed standard deviations ($0.039$, $0.069$, and $0.045$ dB) include variation between mappings. The references vary less on \textsc{garden} and \textsc{counter} ($0.010$--$0.027$ dB) and more on \textsc{kitchen}, where the luminance-only reference has a seed standard deviation of $0.190$ dB, so a pooled noise scale should not be applied per scene. These effects are much smaller than chroma-only failure but establish that chromatic supervision can alter geometry optimization.

The isoluminant-boundary scene is a procedural plate in front of a backdrop, unlit, whose rim is a pure chroma step on its upper half and a chroma-plus-luminance step on its lower half as a within-scene control; luminance across the isoluminant rim stays within $0.11/255$ in every view. RGB, luminance-only, and chroma-only arms were trained with two seeds each. Under the prespecified criterion the difference-in-differences is $+0.028$ against a $0.03$ cutoff, so the failure is not confirmed on that rule; post hoc, luminance-only depth is pulled $19\%$--$30\%$ toward the backdrop across the isoluminant rim against $4.5\%$ across the luminance rim, RGB and chroma-only stay near $2\%$, and chroma-only localizes the rim with about 79k Gaussians. The backdrop side of the rim is unidentifiable for every arm by construction. The ground-truth geometry suite is an analytic room rendered under three material cases (diffuse, colored-specular, and flat-luminance with isoluminant chroma patches), RGB and luminance-only with two seeds each; its prespecified gate, written on all pixels, is failed by RGB itself ($5.6\%$--$7.0\%$ depth error against a $5\%$ gate) because the room walls dominate at grazing incidence. On the objects region, post hoc, luminance-only depth error is $1.00\times$ and $1.02\times$ RGB's on the diffuse and specular cases and $15.6\times$ on the chroma case, where the luminance-only model densifies to about a quarter of RGB's primitives and reaches $22.0$ against $39.9$ dB after $A^*$.

\section*{Acknowledgements}
The authors used Anthropic Claude to assist with drafting and editing the manuscript, writing and reviewing analysis code, and conducting literature and consistency checks. The authors designed and conducted the experiments, verified all results, and take full responsibility for the content.

\bibliographystyle{plainnat}
\bibliography{references}

@article{kerbl20233d,
  title   = {{3D} {Gaussian} Splatting for Real-Time Radiance Field Rendering},
  author  = {Kerbl, Bernhard and Kopanas, Georgios and Leimk{\"u}hler, Thomas and Drettakis, George},
  journal = {ACM Transactions on Graphics},
  volume  = {42},
  number  = {4},
  pages   = {139:1--139:14},
  year    = {2023}
}

@inproceedings{autoopti3dgs2025,
  title     = {From Coarse to Fine: Learnable Discrete Wavelet Transforms for Efficient {3D} {Gaussian} Splatting},
  author    = {Nguyen, Hung and Le, An and Li, Blark Runfa and Nguyen, Truong},
  booktitle = {Proceedings of the IEEE/CVF International Conference on Computer Vision Workshops (ICCVW)},
  pages     = {3170--3179},
  year      = {2025}
}

@article{chao2026palette,
  title   = {Reparametrizing {3D} {Gaussian} Splatting for Real-Time Palette-Based Color and Luminance Editing},
  author  = {Chao, Cheng-Kang Ted and Gingold, Yotam},
  journal = {arXiv preprint arXiv:2609.03897},
  year    = {2026}
}

@inproceedings{cheng2024colorizing,
  title     = {Colorizing Monochromatic Radiance Fields},
  author    = {Cheng, Yean and Wan, Renjie and Weng, Shuchen and Zhu, Chengxuan and Chang, Yakun and Shi, Boxin},
  booktitle = {Proceedings of the AAAI Conference on Artificial Intelligence},
  volume    = {38},
  pages     = {1317--1325},
  year      = {2024}
}

@inproceedings{color3d2026,
  title     = {{Color3D}: Controllable and Consistent {3D} Colorization with Personalized Colorizer},
  author    = {Wan, Yecong and Shao, Mingwen and Wu, Renlong and Zuo, Wangmeng},
  booktitle = {International Conference on Learning Representations (ICLR)},
  year      = {2026}
}

@inproceedings{colmap,
  title     = {{Structure-from-Motion} Revisited},
  author    = {Sch{\"o}nberger, Johannes L. and Frahm, Jan-Michael},
  booktitle = {Proceedings of the IEEE Conference on Computer Vision and Pattern Recognition (CVPR)},
  pages     = {4104--4113},
  year      = {2016}
}

@inproceedings{compact3dgs2024,
  title     = {Compact {3D} {Gaussian} Representation for Radiance Field},
  author    = {Lee, Joo Chan and Rho, Daniel and Sun, Xiangyu and Ko, Jong Hwan and Park, Eunbyung},
  booktitle = {Proceedings of the IEEE/CVF Conference on Computer Vision and Pattern Recognition (CVPR)},
  pages     = {21719--21728},
  year      = {2024}
}

@inproceedings{dhiman2025chromadistill,
  title     = {{ChromaDistill}: Colorizing Monochrome Radiance Fields with Knowledge Distillation},
  author    = {Dhiman, Ankit and Srinath, R and Sarkar, Srinjay and Boregowda, Lokesh R and Babu, R Venkatesh},
  booktitle = {Proceedings of the IEEE/CVF Winter Conference on Applications of Computer Vision (WACV)},
  pages     = {2400--2410},
  year      = {2025}
}

@inproceedings{entropygs2025,
  title     = {{EntropyGS}: An Efficient Entropy Coding on {3D} {Gaussian} Splatting},
  author    = {Huang, Yuning and Pang, Jiahao and Zhu, Fengqing and Tian, Dong},
  booktitle = {Proceedings of the IEEE International Conference on Acoustics, Speech and Signal Processing (ICASSP)},
  pages     = {11732--11736},
  year      = {2026}
}

@article{gsico2026,
  title   = {Structured Image-Based Coding for Efficient {Gaussian} Splatting Compression},
  author  = {Martin, Pedro and Rodrigues, Ant{\'o}nio and Ascenso, Jo{\~a}o and Queluz, Maria Paula},
  journal = {arXiv preprint arXiv:2601.14510},
  year    = {2026}
}

@article{jpeg_chroma_subsampling,
  title   = {The {JPEG} Still Picture Compression Standard},
  author  = {Wallace, Gregory K.},
  journal = {Communications of the ACM},
  volume  = {34},
  number  = {4},
  pages   = {30--44},
  year    = {1991}
}

@article{lcdgs2026,
  title   = {High Dynamic Range {3D} {Gaussian} Splatting via Luminance-Chromaticity Decomposition},
  author  = {Zhang, Kaixuan and Li, Minxian and Ren, Mingwu and Deng, Jiankang and Zhu, Xiatian},
  journal = {arXiv preprint arXiv:2511.12895},
  year    = {2025}
}

@article{lieber2026instant,
  title   = {Instant Colorization of {Gaussian} Splats},
  author  = {Lieber, Daniel and Mock, Alexander and Wandel, Nils},
  journal = {arXiv preprint arXiv:2604.17155},
  note    = {CVPR 2026 Findings track},
  year    = {2026}
}

@inproceedings{luminancegs2025,
  title     = {{Luminance-GS}: Adapting {3D} {Gaussian} Splatting to Challenging Lighting Conditions with View-Adaptive Curve Adjustment},
  author    = {Cui, Ziteng and Chu, Xuangeng and Harada, Tatsuya},
  booktitle = {Proceedings of the IEEE/CVF Conference on Computer Vision and Pattern Recognition (CVPR)},
  pages     = {26472--26482},
  year      = {2025}
}

@inproceedings{niedermayr2024compressed,
  title     = {Compressed {3D} {Gaussian} Splatting for Accelerated Novel View Synthesis},
  author    = {Niedermayr, Simon and Stumpfegger, Josef and Westermann, R{\"u}diger},
  booktitle = {Proceedings of the IEEE/CVF Conference on Computer Vision and Pattern Recognition (CVPR)},
  pages     = {10349--10358},
  year      = {2024}
}

@inproceedings{opti3dgs2025,
  title     = {Optimized {3D} {Gaussian} Splatting Using Coarse-to-Fine Image Frequency Modulation},
  author    = {Farooq, Umar and Guillemaut, Jean-Yves and Thomas, Graham and Hilton, Adrian and Volino, Marco},
  booktitle = {Proceedings of the 22nd ACM SIGGRAPH European Conference on Visual Media Production (CVMP)},
  pages     = {1--10},
  year      = {2025}
}

@article{papantonakis2024memory,
  title   = {Reducing the Memory Footprint of {3D} {Gaussian} Splatting},
  author  = {Papantonakis, Panagiotis and Kopanas, Georgios and Kerbl, Bernhard and Lanvin, Alexandre and Drettakis, George},
  journal = {Proceedings of the ACM on Computer Graphics and Interactive Techniques},
  volume  = {7},
  number  = {1},
  pages   = {1--17},
  year    = {2024}
}

@article{pdfgs2026,
  title   = {{PDF-GS}: Progressive Distractor Filtering for Robust {3D} {Gaussian} Splatting},
  author  = {Seo, Kangmin and Lee, MinKyu and Kim, Tae-Young and Lee, ByeongCheol and An, JoonSeoung and Heo, Jae-Pil},
  journal = {arXiv preprint arXiv:2604.12580},
  year    = {2026}
}

@article{sift,
  title   = {Distinctive Image Features from Scale-Invariant Keypoints},
  author  = {Lowe, David G.},
  journal = {International Journal of Computer Vision},
  volume  = {60},
  number  = {2},
  pages   = {91--110},
  year    = {2004}
}

@misc{sony2026decorrelation,
  title        = {Decorrelation of Spherical Harmonic Coefficients Towards Efficient Compression of {3D} {Gaussian} Splats},
  author       = {Sridhara, Shashank Nelamangala and Zaghetto, Alexandre and Graziosi, Danillo and Tabatabai, Ali},
  howpublished = {U.S. Patent Application Publication No. 2026/0011036 A1},
  note         = {Sony Group Corporation; published January 8, 2026. First circulated as ISO/IEC JTC 1/SC 29/WG 7 input document m69039, Sapporo, July 2024},
  year         = {2026}
}

@article{vangoghgs2026,
  title   = {Thinking Like {Van Gogh}: Structure-Aware Style Transfer via Flow-Guided {3D} {Gaussian} Splatting},
  author  = {Zhou, Lebin and Xiao, Jingchuan and Wang, Zhendong and Wang, Jinhao and Han, Rongduo and Ling, Nam and Ruan, Cihan},
  journal = {arXiv preprint arXiv:2601.10075},
  year    = {2026}
}

@inproceedings{zhang2024fregs,
  title     = {{FreGS}: {3D} {Gaussian} Splatting with Progressive Frequency Regularization},
  author    = {Zhang, Jiahui and Zhan, Fangneng and Xu, Muyu and Lu, Shijian and Xing, Eric},
  booktitle = {Proceedings of the IEEE/CVF Conference on Computer Vision and Pattern Recognition (CVPR)},
  pages     = {21424--21433},
  year      = {2024}
}

@misc{zhou2026structured,
  title        = {Efficient Color Representation for {3D} {Gaussian} Splatting via Structured Spherical Harmonics},
  author       = {Zhou, Qingyang},
  howpublished = {Paper 14217-32, SPIE Optics + Photonics},
  year         = {2026},
  url          = {https://spie.org/optics-photonics/presentation/Efficient-color-representation-for-3D-Gaussian-splatting-via-structured-spherical/14217-32}
}

@article{wang2026surfaces,
  title   = {When {3D} {Gaussian} Splatting Recovers Real Surfaces},
  author  = {Wang, Songhe and Miller, David Johnathan},
  journal = {arXiv preprint arXiv:2608.30054},
  year    = {2026}
}

@article{higs2026,
  title   = {{HiGS}: A Hierarchical Rendering Architecture for Real-Time {3D} {Gaussian} Splatting},
  author  = {Paj{\k{a}}k, Dawid and Bisson, Martin and Lima, Rodolfo},
  journal = {arXiv preprint arXiv:2606.00352},
  year    = {2026}
}

@article{ruderman1998statistics,
  title   = {Statistics of Cone Responses to Natural Images: Implications for Visual Coding},
  author  = {Ruderman, Daniel L. and Cronin, Thomas W. and Chiao, Chuan-Chin},
  journal = {Journal of the Optical Society of America A},
  volume  = {15},
  number  = {8},
  pages   = {2036--2045},
  year    = {1998}
}

@article{mullen1985contrast,
  title   = {The Contrast Sensitivity of Human Colour Vision to Red-Green and Blue-Yellow Chromatic Gratings},
  author  = {Mullen, Kathy T.},
  journal = {The Journal of Physiology},
  volume  = {359},
  number  = {1},
  pages   = {381--400},
  year    = {1985}
}

@inproceedings{rotabulo2024revising,
  title     = {Revising Densification in {Gaussian} Splatting},
  author    = {Rota Bul{\`o}, Samuel and Porzi, Lorenzo and Kontschieder, Peter},
  booktitle = {European Conference on Computer Vision (ECCV)},
  pages     = {347--362},
  year      = {2024}
}

@inproceedings{ye2024absgs,
  title     = {{AbsGS}: Recovering Fine Details in {3D} {Gaussian} Splatting},
  author    = {Ye, Zongxin and Li, Wenyu and Liu, Sidun and Qiao, Peng and Dou, Yong},
  booktitle = {Proceedings of the 32nd ACM International Conference on Multimedia (MM)},
  pages     = {1053--1061},
  year      = {2024}
}

@inproceedings{mallick2024taming,
  title     = {Taming {3DGS}: High-Quality Radiance Fields with Limited Resources},
  author    = {Mallick, Saswat Subhajyoti and Goel, Rahul and Kerbl, Bernhard and Carrasco, Francisco Vicente and Steinberger, Markus and De La Torre, Fernando},
  booktitle = {SIGGRAPH Asia 2024 Conference Papers},
  pages     = {1--11},
  year      = {2024}
}

@inproceedings{barron2022mipnerf360,
  title     = {{Mip-NeRF} 360: Unbounded Anti-Aliased Neural Radiance Fields},
  author    = {Barron, Jonathan T. and Mildenhall, Ben and Verbin, Dor and Srinivasan, Pratul P. and Hedman, Peter},
  booktitle = {Proceedings of the IEEE/CVF Conference on Computer Vision and Pattern Recognition (CVPR)},
  pages     = {5470--5479},
  year      = {2022}
}

@article{knapitsch2017tanks,
  title   = {Tanks and Temples: Benchmarking Large-Scale Scene Reconstruction},
  author  = {Knapitsch, Arno and Park, Jaesik and Zhou, Qian-Yi and Koltun, Vladlen},
  journal = {ACM Transactions on Graphics},
  volume  = {36},
  number  = {4},
  pages   = {78:1--78:13},
  year    = {2017}
}

@article{hedman2018deep,
  title   = {Deep Blending for Free-Viewpoint Image-Based Rendering},
  author  = {Hedman, Peter and Philip, Julien and Price, True and Frahm, Jan-Michael and Drettakis, George and Brostow, Gabriel},
  journal = {ACM Transactions on Graphics},
  volume  = {37},
  number  = {6},
  pages   = {257:1--257:15},
  year    = {2018}
}

@inproceedings{verbin2022refnerf,
  title     = {{Ref-NeRF}: Structured View-Dependent Appearance for Neural Radiance Fields},
  author    = {Verbin, Dor and Hedman, Peter and Mildenhall, Ben and Zickler, Todd and Barron, Jonathan T. and Srinivasan, Pratul P.},
  booktitle = {Proceedings of the IEEE/CVF Conference on Computer Vision and Pattern Recognition (CVPR)},
  pages     = {5481--5490},
  year      = {2022}
}

@article{wang2004ssim,
  title   = {Image Quality Assessment: From Error Visibility to Structural Similarity},
  author  = {Wang, Zhou and Bovik, Alan C. and Sheikh, Hamid R. and Simoncelli, Eero P.},
  journal = {IEEE Transactions on Image Processing},
  volume  = {13},
  number  = {4},
  pages   = {600--612},
  year    = {2004}
}

@inproceedings{zhang2018lpips,
  title     = {The Unreasonable Effectiveness of Deep Features as a Perceptual Metric},
  author    = {Zhang, Richard and Isola, Phillip and Efros, Alexei A. and Shechtman, Eli and Wang, Oliver},
  booktitle = {Proceedings of the IEEE/CVF Conference on Computer Vision and Pattern Recognition (CVPR)},
  pages     = {586--595},
  year      = {2018}
}

@inproceedings{ren2026fastgs,
  title     = {{FastGS}: Training {3D} {Gaussian} Splatting in 100 Seconds},
  author    = {Ren, Shiwei and Wen, Tianci and Fang, Yongchun and Lu, Biao},
  booktitle = {Proceedings of the IEEE/CVF Conference on Computer Vision and Pattern Recognition (CVPR)},
  pages     = {26094--26103},
  year      = {2026}
}

\end{document}